%% file: paper.tex
\documentclass[]{fairmeta}

\usepackage{newtxtext,newtxmath} 

\usepackage{graphicx}
\usepackage{adjustbox}

\usepackage{booktabs}
\usepackage{multirow}
\usepackage{tabularx,makecell}
\usepackage[table]{xcolor}
\newcolumntype{Y}{>{\centering\arraybackslash}X}

\usepackage{siunitx}
\usepackage{threeparttable}

\usepackage{pifont}

\newcommand{\xmark}{\ding{55}}

\usepackage{wrapfig}   
\usepackage{caption}
\usepackage{algorithm}
\usepackage{algpseudocode} 

\usepackage{enumitem}

\title{Unifying Contrastive and Generative Objectives for Visual Understanding and Text-to-Image Generation}

\author[1,*]{Chao Li}
\author[1]{Tianhong Li}
\author[2]{Sai Vidyaranya Nuthalapati}
\author[2]{Hong-You Chen}
\author[2]{Satya Narayan Shukla}
\author[2]{Jianpeng Cheng}
\author[2]{Yonghuan Yang}
\author[2]{Jun Xiao}
\author[2]{Xiangjun Fan}
\author[2]{Aashu Singh}
\author[1]{Dina Katabi}
\author[2]{Shlok Kumar Mishra}

\affiliation[1]{MIT Computer Science \& Artificial Intelligence Laboratory}
\affiliation[2]{Meta AI}

\contribution[*]{Work done at Meta}

\abstract{\input{sections/abstract}}

\date{\today}
\correspondence{Chao Li at \email{chaoli@mit.edu}}

\metadata[Code]{\url{https://github.com/chaoli-charlie/dream}}

\begin{document}

\maketitle
\input{sections/introduction}
\input{sections/related_works}
\input{sections/implementation}
\input{sections/methods}
\input{sections/results}
\input{sections/discussion}

\clearpage
\newpage
\bibliographystyle{assets/plainnat}
\bibliography{paper}

\clearpage
\newpage
\beginappendix

\input{sections/appendix}

\end{document}

%% file: sections/abstract.tex
Unifying text-image contrastive learning and text-to-image (T2I) generation in a single end-to-end model is challenging because the two objectives demand opposing masking regimes: contrastive alignment needs near-complete visible tokens, while masked generative modeling needs heavy corruption. 
We introduce DREAM, a unified framework that resolves this conflict through \textit{Masking Warmup}, a schedule that shifts the center of the masking distribution over training, so low and high masking ratios coexist at every step. This co-exposure lets a single jointly-trained encoder serve both objectives. The resulting stable optimization unlocks \textit{Semantically Aligned Decoding} at inference: the text encoder, trained against visual embeddings at all masking ratios, can score partially generated images and select the best trajectory with as little as 12.5\% of the image decoded, improving both FID and throughput. 
DREAM outperforms its single-objective baselines, CLIP and FLUID: on ImageNet linear-probing (+1.1\%), 5-shot transfer (+4.1\%), ADE20K segmentation (+1.9\%), and NYU depth estimation (+6.25\%) over CLIP, and on CC12M FID (+6.2\%) over FLUID while maintaining CLIP Score. Together, these gains show that text-image contrastive and generative objectives, when properly unified, are synergistic rather than competing.

%% file: sections/introduction.tex
\section{Introduction}
Multimodal learning has long been split between \textbf{models that represent} and \textbf{models that generate}. Discriminative vision--language systems such as CLIP \cite{radford2021clip} learn semantically rich features through contrastive alignment, while text-to-image (T2I) generative models, including diffusion-based \cite{ramesh2022hierarchical, saharia2022imagen} and masked autoregressive (MAR) approaches \cite{li2024autoregressive, fan2024Fluid}, learn conditional pixel distributions through aggressive corruption and reconstruction. Unifying them inside a single, end-to-end trainable network has proved persistently difficult \cite{chen2023vlp, wu2025harmonizing}. Existing unified models sidestep rather than resolve this difficulty by freezing one pathway \cite{zheng2025diffusion}, relying on external teachers \cite{yu2025repa}, or decoupling the two objectives across separate training stages.

The root cause is structural: the two losses require contradictory masking regimes. Contrastive alignment needs the visible token set to carry near-complete semantics, since computing InfoNCE over heavily masked images degrades text-aligned structure \cite{li2023scaling}; masked generative modeling needs the opposite, since high masking ratios are necessary for the decoder to learn a useful conditional distribution over latent tokens \cite{li2024autoregressive}. In our controlled single-stage, unfrozen-encoder setting, the conflict is severe: naive joint training collapses linear probing accuracy to \textbf{4.6\%} while also failing to converge on generation.

Because the conflict originates in the masking-ratio requirements of the two objectives, the tension should be addressable at the level of the masking distribution itself, rather than through architectural workarounds that avoid co-optimization. We introduce \textbf{DREAM}, a unified framework built on this insight through \textbf{Masking Warmup}, where the center of the masking distribution shifts over training. Early on, the distribution is centered low, letting the encoder establish a stable, text-aligned representation under the regime contrastive alignment requires. The center then shifts toward the high-mask regime the decoder needs, so these warmed-up features anchor the generative objective rather than being overwritten by it. Crucially, the distribution stays wide throughout: low and high masking ratios coexist at every step, so the encoder continues text alignment even as the decoder receives heavy corruption. This sustained co-exposure is what allows both representation and generation quality to keep improving after the schedule terminates (Fig.~\ref{fig:stability}).

\begin{wrapfigure}{l}{0.5\textwidth}
  \centering
  \includegraphics[width=\linewidth]{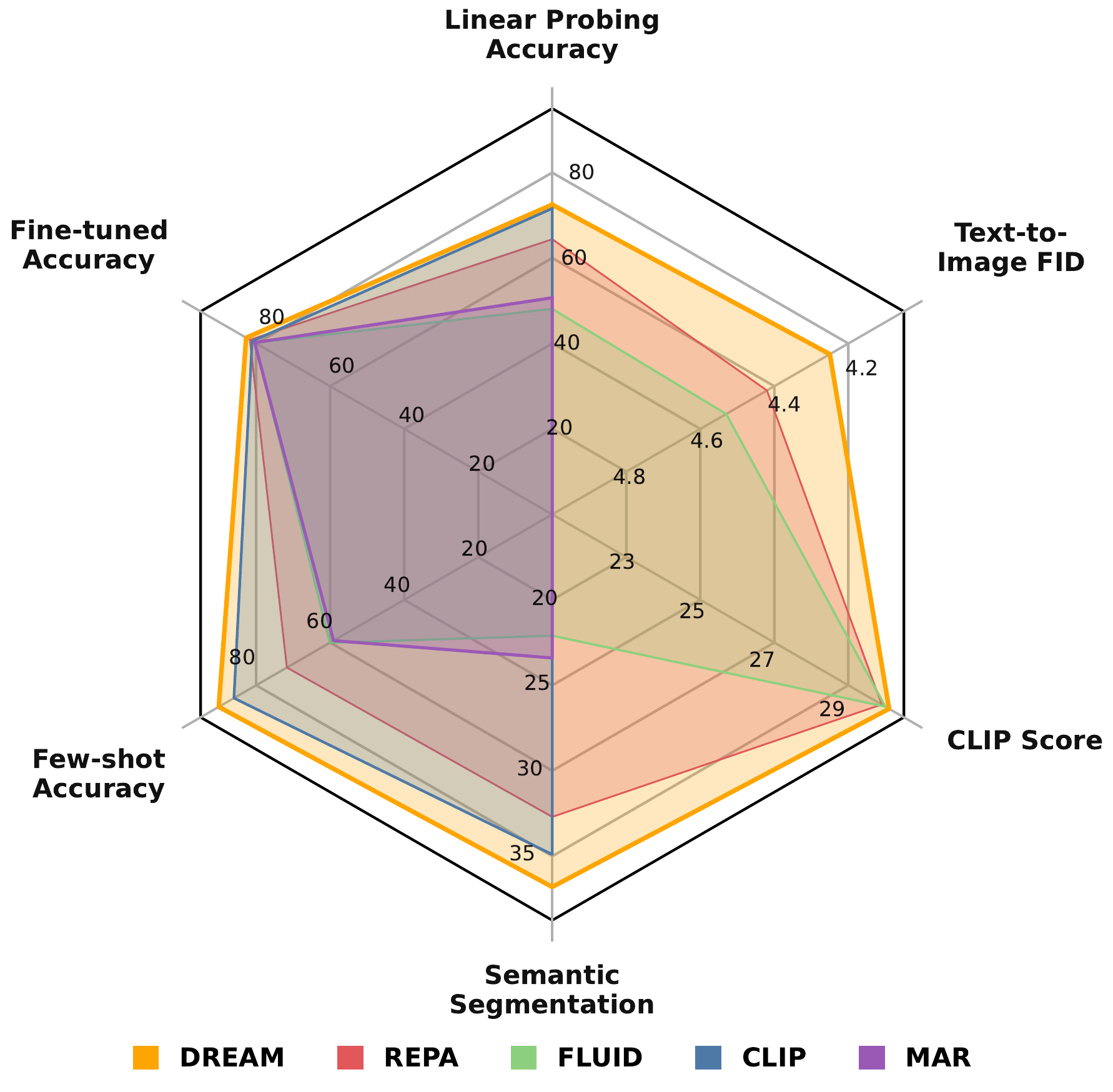}
  \caption{\textbf{Comparison of representation and T2I generation models on CC12M.} DREAM (yellow) outperforms across all discriminative and generative axes, unifying visual understanding with T2I generation in a single stage model.}
  \label{fig:summary}
  \vspace{-1em}
\end{wrapfigure}

Joint optimization is not merely a training convenience; it unlocks an inference-time capability that decoupled models structurally cannot replicate. Following MAR~\cite{li2024autoregressive}, DREAM's encoder drops masked positions and operates on sparse subset of visible tokens. Because the text encoder is jointly trained against these visual embeddings across the full range of visibility levels, it learns to align text with images that are only partially decoded. We exploit this with \textbf{Semantically Aligned Decoding (SD)}: with as little as 12.5\% of the image generated, the text encoder can discriminate between candidate trajectories and select the one most aligned with the prompt. The improvement in generation FID by 7.0\% through SD is evidence of the synergy between text-image contrastive and generative loss enabled through joint optimization. External CLIP rerankers~\cite{ramesh2022hierarchical, chang2023muse} trained only on fully visible images cannot reliably score such sparse embeddings, leading to overhead in generating the full image before selection. SD effectiveness and efficiency is reflected through superior performance in both FID (+5.6\%) and throughput (+10.1\%) over external reranking models.

Fig.~\ref{fig:summary} summarizes the gains. DREAM improves discriminative representations over CLIP, the strongest representation-aligned baseline: $+1.1\%$ linear probing on ImageNet-1K, $+4.1\%$ in 5-way 5-shot transfer, $+1.9\%$ on ADE20K, and $6.25\%$ lower RMSE on NYU Depth v2. On generation, DREAM improves T2I FID by 6.2\% over FLUID on CC12M while maintaining competitive CLIP score, and on zero-shot MS-COCO it achieves the highest CLIP score among all baselines. The pattern is itself informative: gains on dense prediction trace to the generative reconstruction loss, which encourages pixel-aligned features \cite{he2022mae}, while gains on classification and retrieval trace to the contrastive loss. We summarize our contributions as follows:

\begin{itemize}[noitemsep, topsep=0pt]
    \item \textbf{Diagnosing the Unification Conflict at Its Source.} We identify the root cause of unstable joint training as a mismatch in the masking-ratio requirements of the two objectives.

    \item \textbf{Masking Warmup for Stable Single-Stage Joint Training.} We introduce \emph{Masking Warmup}, a schedule on the masking distribution that resolves this conflict and, to our knowledge, enables the first stable single-stage end-to-end joint training of text-image contrastive and T2I generative objectives with a single architecture.

    \item \textbf{Discriminative and Generative Objectives are Synergistic.} Leveraging this stable training regime, we show through extensive evaluation on classical discriminative and generation benchmarks that the two objectives are mutually reinforcing rather than competing.

    \item \textbf{Semantically Aligned Decoding.} We introduce an efficient and effective self-guided inference strategy that scores partially generated images using the model's own contrastive representations, a capability uniquely enabled only from unifying the two objectives.
\end{itemize}

%% file: sections/related_works.tex
\section{Related Work}

\input{plots/training_plot}

\paragraph{Unifying Representation and T2I Generation.}
Prior efforts fall into three fronts.
\textit{Joint generative modeling without contrastive alignment.} Transfusion~\cite{zhou2024transfusion} jointly trains next-token prediction on text and diffusion on images inside one transformer and 4M~\cite{mizrahi20234m} performs masked token prediction across multiple modalities. These models unify generation across modalities but include no contrastive alignment objective, so the contrastive–generative conflict we address does not arise.
\textit{Multi-stage pipelines avoid co-optimization:} Show-o~\cite{xie2024showo}, Janus~\cite{wu2025janus}, and Harmon~\cite{wu2025harmonizing} split training into multiple stages. Janus also uses separate understanding and generation encoders. 
\textit{Single-stage with intra-modal contrast:} MAGE~\cite{li2023mage} and ST-AR~\cite{yue2025understand} add image–image contrastive losses to unconditional and class-conditional generation. Because their contrast is intra-modal and neither conditions on text, the text-image conflict we address does not arise.
\textit{Representation alignment to a frozen external encoder:}  REPA~\cite{yu2025repa} regularizes diffusion-transformer features to match a frozen pretrained encoder; RAE~\cite{zheng2025diffusion} runs diffusion directly in the latent space of a frozen pretrained encoder. In both, the alignment target is a fixed external vision model rather than a jointly-trained text encoder. 

\vspace{-1em}

\paragraph{Inference Time Decoding}
Scoring partial generations to prune unpromising trajectories early has emerged as an effective alternative to decoding all candidates to completion, instantiated by future discriminators in language modeling~\cite{yang-klein-2021-fudge}, hand-designed early-timestep scores for image editing~\cite{kim2025early}, and auxiliary selectors for 3D Gaussian diffusion~\cite{yin2025trim}. In T2I generation, the common practice is post-hoc CLIP reranking~\cite{chang2023muse, ramesh2022hierarchical}, which scores fully decoded candidates with an external pretrained model. All of these approaches rely on a discriminator external to the generator, and none provides a model-internal, text-aligned scorer for text-to-image generation.

%% file: plots/training_plot.tex
\begin{figure*}[t]
    \centering
    \includegraphics[width=\linewidth]{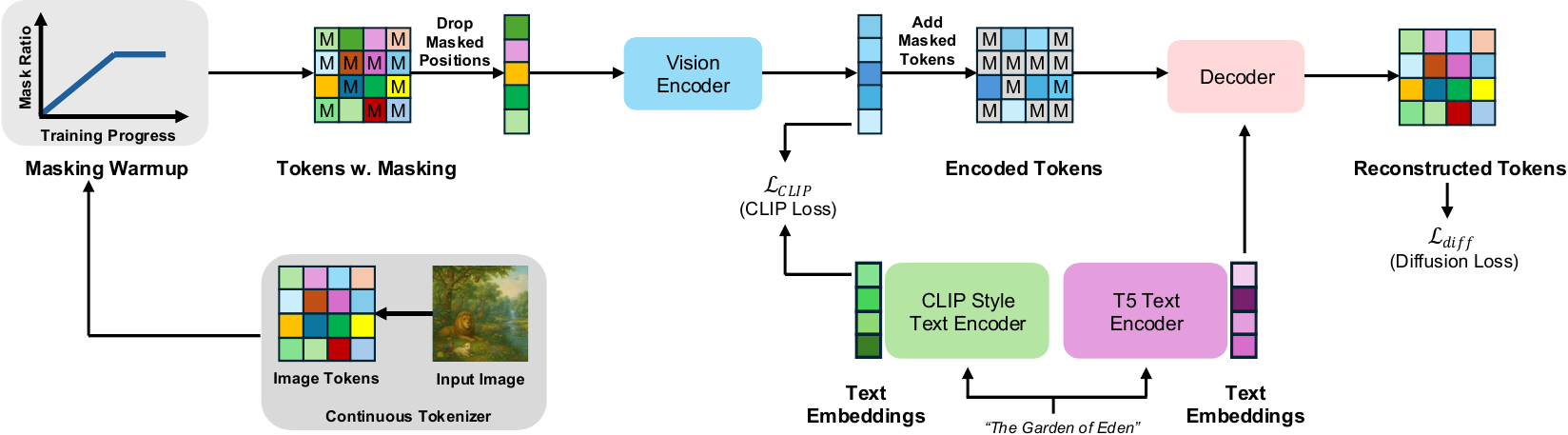}
    \caption{\textbf{DREAM training framework.} 
    Images are encoded into continuous tokens via Stable Diffusion VAE and randomly masked and dropped following a masking warmup schedule. The vision encoder is trained contrastively with text, and the decoder conditions on text to predict masked tokens with a diffusion-based reconstructive loss. Text conditioning is introduced only in the decoder, ensuring the encoder learns visual representations without a text shortcut.
    }
    \label{fig:pretraining}
\end{figure*}

%% file: sections/implementation.tex
\section{Implementation Details}
\label{sec:implementation}

\paragraph{Data.} We train on Conceptual 12M (CC12M)~\cite{changpinyo2021cc12m}, containing 11.3M text-image pairs. Images are center-cropped to $256 \times 256$ with horizontal flipping as augmentation~\cite{li2024autoregressive}. We evaluate representations on ImageNet-1K~\cite{deng2009imagenet} and generation quality on CC12M and MS-COCO~\cite{lin2014microsoft}.

\vspace{-1em}

\paragraph{Training.} All models are trained for 49 epochs using AdamW~\cite{loshchilov2017decoupled} optimizer ($\beta_1 = 0.9, \beta_2 = 0.95$) with batch size 2048 and constant learning rate with 12-epoch linear warmup to a maximum of $8\times10^{-4}$. DREAM additionally employs the 36-epoch progressive masking warmup described above. Evaluation uses the exponential moving average (EMA) of model weights with a decay rate of 0.9999.

\vspace{-1em}

\paragraph{Masking Warmup.} The masking ratio is sampled from a truncated Gaussian distribution with a fixed standard deviation. The mean of the distribution increases linearly from 0 to 1.0 over the first 36 epochs. After that point, the mean is fixed at 1.0, corresponding to a fully masked image.

%% file: sections/methods.tex
\section{Method}
DREAM adopts a ViT-based encoder–decoder architecture over continuous image latents. The encoder learns language-aligned visual features, and the decoder generates images conditioned on text through a diffusion based reconstruction loss. Importantly, text conditioning is applied only in the decoder, ensuring the encoder learns visual representations without relying on language shortcuts (see Fig.~\ref{fig:pretraining}).

\subsection{Architecture}
\paragraph{Continuous Tokenization.}
We encode  images into continuous latent representations using the pretrained VAE encoder from Stable Diffusion~\cite{rombach2022high}.
This continuous tokenization preserves fine-grained spatial information while maintaining computational efficiency.

\vspace{-1em}

\paragraph{Vision Encoder.}
Our encoder follows MAR's architecture~\cite{li2024autoregressive}. We prepend learnable buffer tokens to the encoder input to enhance representation capacity and training stability. The encoder processes only the buffer tokens and the unmasked tokens.

\vspace{-1em}

\paragraph{Text Encoders.}
We employ two text encoders with distinct roles in training.
For \textbf{contrastive alignment}, captions are tokenized using the OpenAI CLIP tokenizer~\cite{radford2021clip} (77 tokens) and encoded by a CLIP text transformer following~\cite{tian2023stablerep}, projecting the text features into the vision encoder’s latent space.
For \textbf{generation}, captions are tokenized with SentencePiece~\cite{kudo2018sentencepiece} (128 tokens) and encoded by a frozen T5-XXL~\cite{raffel2020exploring}.
A lightweight 6-layer text aligner then projects the T5 embeddings into the decoder’s latent space, providing conditioning for masked autoregressive generation.

\vspace{-1em}

\paragraph{Text-to-Image Decoder.}
Our decoder adopts FLUID’s architecture but is paired with the encoder described above, forming a unified encoder–decoder model.
Each block comprises self-attention, cross-attention, and feed-forward layers.
Self-attention operates on visual tokens, while cross-attention conditions generation on text features.
During training, masked visual tokens are replaced with a learnable token, and the decoder predicts these masked tokens using bidirectional attention, similar to BERT. 
The encoder and decoder each contain half of the total transformer blocks.
Our Large (L) configuration uses 32 layers (16 per module) with a hidden width of 1024, totaling 570M parameters.

\vspace{-1em}

\paragraph{Output Head.}
To predict continuous latent tokens, we attach a lightweight six-layer MLP diffusion head \cite{li2024autoregressive} that matches the transformer’s embedding dimension. The diffusion process follows the improved DDPM formulation \cite{nichol2021improved, li2024autoregressive}, using a cosine noise schedule with 1000 steps during training and 100 resampled steps during inference.

\input{plots/semantic_decoding_plot}

\subsection{Training Objectives}
\label{training_objectives}
\paragraph{Diffusion Reconstruction Loss}
We adopt the same diffusion loss used in \cite{li2024autoregressive, fan2024Fluid}, where the conditional distribution $p(x\mid z)$ via denoising is modelled as follows:

\begin{equation}
\mathcal{L}_{\text{diff}}(z, x) = \mathbb{E}_{\epsilon, t} \left[||\epsilon - \epsilon_\theta(x_t|t, z)||^2\right]
\end{equation}

where $\epsilon \sim \mathcal{N}(0, I)$ is Gaussian noise, and $x_t = \sqrt{\bar{\alpha}_t}x + \sqrt{1 - \bar{\alpha}_t}\epsilon$ represents the noisy latent at timestep $t$ under variance schedule $\bar{\alpha}_t$. The denoising network $\epsilon_\theta$ predicts noise conditioned on encoder embeddings $z$, enabling end-to-end gradient. We sample four independent noise levels per image to improve gradient estimates without recomputing $z$.

We compute diffusion loss for samples that have greater than 50\% of its tokens masked, since high masking ratios are essential for effective generative modeling \cite{li2024autoregressive}.

\vspace{-1em}

\paragraph{Contrastive Learning}
We further incorporate CLIP-style contrastive learning \cite{radford2021clip} to align image and text representations.
Given $N$ paired samples $(x_i^I, x_i^T)$, we compute the InfoNCE loss as:

\small
\begin{equation}
\mathcal{L}_{\text{I}} =
- \sum_{i=1}^{N}
\log
\frac{
e^{\,\text{sim}(f_I(\text{aug}_I(x_i^I)), f_T(x_i^T)) / \tau}
}{
\sum_{k=1}^{N}
e^{\,\text{sim}(f_I(\text{aug}_I(x_i^I)), f_T(x_k^T)) / \tau}
}.
\label{eq:clip_loss}
\end{equation}
\normalsize

where $\text{sim}(\cdot, \cdot)$ denotes cosine similarity and $\tau$ is a learnable temperature parameter.
A symmetric text-to-image term $\mathcal{L}_T$ is computed analogously, and the final contrastive objective is $\mathcal{L}_{\text{clip}} = (\mathcal{L}_I + \mathcal{L}_T)/2$.

We cap the maximum masking ratio at 75\% for CLIP loss computation, as excessive masking degrades discriminative learning ~\cite{li2023scaling, li2023mage}.

\paragraph{Joint Optimization}
The total loss jointly optimizes generation and alignment:

\begin{equation}
\mathcal{L} = \mathcal{L}_{\text{diff}} + \lambda \cdot \mathcal{L}_{\text{clip}},
\end{equation}

where $\lambda$ balances the diffusion and contrastive terms.

\subsection{Inference}
\label{sec:inference}
\paragraph{Representation Extraction.}
For linear probing and finetuning, encoder output features are globally average-pooled and fed into task-specific heads (linear probes or fine-tuned classifiers). For semantic segmentation and depth estimation, we freeze the vision encoder and evaluate features and follow the linear-probe protocol of \cite{simeoni2025dinov3}.

\vspace{-1em}

\paragraph{Image Generation.}
For image generation, DREAM follows MAR’s next set-of-tokens prediction strategy~\cite{li2024autoregressive}: starting from a fully masked sequence, the masking ratio is annealed from 1.0 to 0 using a cosine schedule~\cite{chang2022maskgit, li2023mage} over 64 steps by default. Decoding uses temperature sampling and fully randomized token orders at every step.


\subsection{Semantically Aligned Decoding.} 
We introduce a zero-shot semantically aligned decoding strategy that selects the most text-aligned candidate during the decoding process.
As illustrated in Fig.~\ref{fig:inference}, given a prompt, DREAM spawns $K$ candidates—independent decoding trajectories initialized by different stochastic seeds. After a small fraction of the 64 decoding steps (t$<<$64), DREAM produces partially decoded latents for all $K$ candidates. These partial latents are fed to the vision encoder to obtain visual representations, which are scored for text–image alignment against the prompt using DREAM’s contrastive text encoder (shown to reliably retrieve partially decoded images, see Appendix~\ref{sec:zeroshot_results}). Only the top-scoring candidate continue for the remaining decoding steps to reach the full image.

This approach differs from prior image selection methods~\cite{chang2023muse, ramesh2022hierarchical} in three ways: (1) selection occurs at the latent level; (2) only a fraction of the image is decoded, avoiding the cost of generating complete images; (3) the model uses its own learned alignment rather than external vision-language models.

%% file: plots/semantic_decoding_plot.tex
\begin{figure*}[t]
    \centering
    \includegraphics[width=0.9\linewidth]{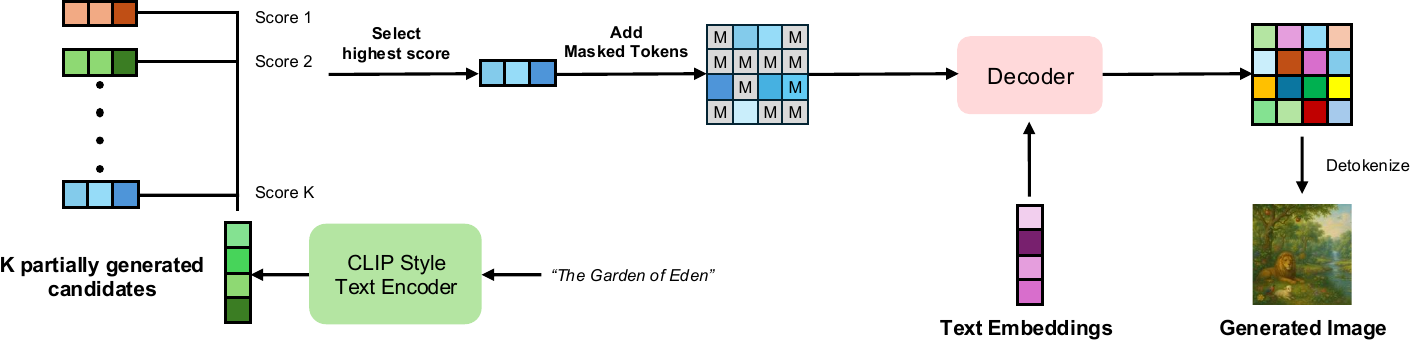}
    \caption{\textbf{Semantically Aligned Decoding.} 
    The model spawns $K$ parallel candidates, each partially decoded to an intermediate timestep $t$. The encoder scores each candidate by comparing its visual embedding to the prompt embedding, and only the top-scoring candidate is fully decoded.
    }
    \label{fig:inference}
\end{figure*}

%% file: sections/results.tex
\section{Experiments}
\label{results}

We evaluate DREAM on representation quality, measured by Linear Probing accuracy on ImageNet-1K \cite{deng2009imagenet}, and text-to-image generation quality, measured by Fréchet inception distance (FID) on CC12M \cite{changpinyo2021cc12m}. All models are trained for 49 epochs on CC12M.

\subsection{Masking Warmup Resolves the Contrastive--Generative Conflict}
\label{sec:masking_conflict}
We first investigate how the masking distribution governs the joint optimization of the text-image contrastive and generative objectives. Across all variants, we hold architecture, training budget, and contrastive loss weight ($\lambda = 0.005$) fixed, varying only the per-step masking distribution. For this controlled study, we set $\sigma = 0.45$, which yields stable convergence (refer to Appendix~\ref{sec:masking_stability}).

\vspace{-1em}

\paragraph{Finding 1: Naive joint training collapses under fixed high masking.}
We consider a naive co-optimization implementation, \textbf{FX}, which follows FLUID's~\cite{fan2024Fluid} fixed truncated Gaussian centered at a $1.0$ masking ratio. As shown in Table~\ref{tab:ablate_schedule}, this method collapses the representation space compared to the single objective models (CLIP and FLUID), and yields poor performance on both representation and generation. \textit{Under sustained high masking, the visible token set carries too little global semantics to be aligned and contrasted with text}. Hence the encoder--decoder pair fails to converge.

\vspace{-1em}

\paragraph{Finding 2: Stable schedules trade generation for representation.}
We compare two alternatives: \textbf{CD}, a cooldown schedule whose mean linearly decreases from $1.0$ to $0.0$ over the first 36 epochs; and \textbf{WM} a warmup schedule whose mean linearly increases from $0.0$ to $1.0$ over the same horizon and is held fixed afterward. \textbf{CD} and \textbf{WM} share an identical distributional shape, duration, differing \emph{only} in the direction of the schedule.

\input{results/masking_schedule}

While \textbf{CD} outperform CLIP on Linear Probing, it underperforms the FLUID baseline on FID (5.80 vs 4.53). \textbf{CD} concentrates high-mask exposure early, before the encoder has established a stable text-aligned representation. It ends in a low-mask regime that provides near-zero generative signal, thus sacrificing generation quality for representation gains.

\vspace{-1em}

\paragraph{Finding 3: Masking Warmup is synergistic.}
\textbf{WM} demonstrates that stable co-optimization is not only possible but synergistic: it surpasses both single-objective baselines (CLIP for representation, FLUID for generation) within a single jointly trained model. The contrast with \textbf{CD} is particularly informative---because the two schedules differ only in direction, the gap between them isolates the importance of \emph{progression}: the contrastive objective must establish a stable semantic basin under low masking \emph{before} the high-mask generative signal arrives. Crucially, the distribution remains wide ($\sigma = 0.45$) at every step, so low- and high-mask samples coexist throughout; this preserves stable joint optimization even after warmup ends and the mean is held high (Fig.~\ref{fig:stability}), since the contrastive objective continues to receive sufficiently visible image context.

In Appendix~\ref{two_stage}, we further distinguish Masking Warmup from explicit staged optimization: a two-stage baseline that pretrains the encoder with CLIP and then trains the generative objective with an unfrozen encoder exhibits a similar collapse regardless of stage-1 duration, confirming the failure is structural and resolved only by sustained contrastive signal during high-masking generative training.

\vspace{-1em}

\subsection{Effect of Diffusion Reconstruction}
\label{sec:mar_effect}
Next, we show that DREAM's Linear Probing gains are not from the masking warmup schedule alone, but from the synergy between contrastive and generative objectives that the warmup enables. To isolate this, we compare DREAM against an ablation that removes the diffusion reconstruction loss, leaving only the contrastive objective trained under the same warmup schedule.

As shown in Table~\ref{tab:ablate_schedule}, removing the reconstruction loss \emph{degrades} Linear Probing from $72.5\%$ to $70.8\%$---below even the CLIP baseline ($71.6\%$). Without a complementary objective in the high-mask regime, the visible token set carries too little global semantics for InfoNCE to compute meaningful alignments, and the encoder's representations deteriorate as training progresses. DREAM avoids this collapse: during the high masking regime at the end of Masking Warmup, the diffusion reconstruction loss provides spatial supervision precisely where InfoNCE's signal weakens, while contrastive alignment anchors the encoder against the representational drift that pure generative training induces.

\vspace{-1em}

\subsection{Unified Performance}
\label{sec:unified}

\input{results/classification}

We now show the synergistic relationship enabled by Masking Warmup evaluated across classical discriminative representation and generation metrics. Our comparisons target DREAM's problem formulation: single-stage, end-to-end joint training with an unfrozen encoder. Within this formulation, we compare three baseline families sharing the same encoder–decoder architecture and differing only in objective: (1) contrastive-only (CLIP); (2) generative-only (MAR, reimplemented without class conditioning; FLUID, adapted from decoder-only to encoder–decoder by adding text conditioning to the decoder); and (3) representation-aligned generative (REPA), which extends FLUID by aligning unmasked tokens at encoder layer 6 with a pretrained CLIP-L encoder~\cite{radford2021clip}; originally proposed for class-conditioned diffusion~\cite{yu2025repa}, we adapt it with text conditioning and use CLIP-L instead of DINO-v2, which yields stronger T2I fidelity (Appendix~\ref{sec:repa_teacher}). DREAM applies Masking Warmup to jointly train FLUID's generative objective with a CLIP contrastive loss on the encoder's final layer. In these unified results, we report DREAM trained with $\sigma = 0.55$ and provide ablations in Appendix~\ref{sec:std_mask}.

\subsubsection{Discriminative Representations}
\label{sec:rep_learning}

\paragraph{Linear Probing.}
Linear probing on frozen features reflects representation quality directly. As shown in Table~\ref{tab:unified_baselines}, generative models, both unconditional (MAR) and text-conditioned (FLUID), learn substantially weaker representations than CLIP, while CLIP-aligned models (REPA, DREAM) close this gap. DREAM attains 72.7\% accuracy, surpassing FLUID by 24.6\% and exceeding CLIP by 1.1\% despite sharing the same contrastive objective, indicating that coupling language-aligned contrastive learning with diffusion-based reconstruction yields more transferable representations. Our CLIP baseline on SD-VAE tokens matches the linear probing accuracy of pixel-based CLIP reported in StableRep~\cite{tian2023stablerep}, confirming the tokenization does not degrade representation quality.

\vspace{-1em}

\paragraph{Fine-tuning.}
Following \cite{li2023mage}, we fine-tune on ImageNet-1K and evaluate on out-of-domain ImageNet variants (Table~\ref{tab:unified_baselines}). DREAM achieves the strongest in-domain accuracy, surpassing CLIP by $+1.6\%$ and REPA by $+1.0\%$ on ImageNet-1K, and the highest out-of-domain accuracy, exceeding CLIP by $+2.8\%$ and REPA by $+2.4\%$ on average. CLIP-aligned models (REPA, DREAM) reliably outperform both CLIP and their generative-only counterparts after fine-tuning, confirming that integrating a generative objective is synergistic with learning transferable features. DREAM's gains are especially pronounced on the hardest variants (IN-A, IN-H), underscoring its out-of-distribution generalization.

\vspace{-1em}

\paragraph{Few-shot Learning.}
Prior work has established that high-quality representations are critical for few-shot image classification~\cite{wang2019simpleshot, tian2020rethinking, dhillon2019baseline}. As shown in Appendix Table~\ref{tab:fewshot}, DREAM consistently outperforms all baselines, exceeding the next-best model (CLIP) by $+4.1\%$ on average.

\vspace{-1em}

\paragraph{Dense Prediction.}
Appendix Table~\ref{tab:segmentation_depth_combined} shows the result on dense prediction. DREAM achieves 36.8\% mIoU on ADE20K ($+1.9\%$ over CLIP) and 0.60 RMSE on NYU Depth v2, matching REPA while outperforming CLIP by 6.25\%. These gains indicate that the diffusion reconstruction objective enhances spatial grounding by encouraging pixel-aligned features that transfer to dense prediction.

\vspace{-1em}

\paragraph{Zero-shot.}
Since DREAM is trained with variable masking ratios, it is more resilient on a sparse subset of visible tokens. We compare CLIP, CLIP with masking warm-up (CLIP-M), and DREAM on zero-shot accuracy across masking levels (Appendix~\ref{sec:zeroshot_results}). Masking Warmup improves robustness for both CLIP-M and DREAM over CLIP, with gains once $>$20\% of the image is dropped. DREAM consistently outperforms CLIP-M at all levels, reflecting the benefit of locally grounded representations from diffusion-based reconstruction. At high masking ($>0.8$), DREAM achieves over 6.2$\times$ the zero-shot accuracy of CLIP, highlighting the resilience from joint contrastive–generative training.

\input{results/unified_results}

\subsubsection{T2I Generation}
\label{sec:t2i_performance}
We evaluate DREAM on CC12M and zero-shot MS-COCO using FID and CLIP Score (CS), following \citet{chang2023muse}. As shown in Table~\ref{tab:unified_baselines}, DREAM achieves the best FID and CS on CC12M, reducing FLUID's FID by 6.2\% while slightly improving CS, indicating that coupling contrastive alignment with generation enhances reconstruction fidelity without compromising semantic alignment. \textsc{REPA} also improves over FLUID (FID 4.42) but trails DREAM by 4.0\%. On zero-shot MS-COCO, DREAM attains the highest CS (31.5) but a slightly worse FID than FLUID (10.4 vs. 9.62), a tradeoff also observed for REPA and consistent with representation-aligned methods being more sensitive to caption-distribution shift.

\textbf{Semantically Aligned Decoding improves both fidelity and alignment.} We isolate the contribution of SD by comparing DREAM with and without it under a fixed decoding budget ($T = 64$). SD improves FID by 7.0\% and CS by 3.4\% on CC12M, and CS by 3.3\% on MS-COCO at comparable FID. This is direct evidence of synergy: the jointly trained contrastive encoder selects better decoding trajectories for the same generative process, actively benefiting generation rather than merely coexisting with it. Appendix~\ref{sec:diversity} shows SD does not reduce diversity relative to external CLIP reranking.


\subsubsection{Efficiency Analysis}
\label{sec:inference_efficiency}

\input{results/efficiency}

Table~\ref{tab:dream_k} reports throughput, FID, and CLIP score under matched NFE budgets. The benefit of SD decomposes into two parts:

\textit{First, scoring with the internal text encoder beats external CLIP reranking even without partial-latent selection.} At matched NFE=128, DREAM with $K=1$ and $T=128$ reaches FID 4.34, lower than DREAM with $K=2$ reranked by an external CLIP (4.50). Spending the budget on more decoding steps for a single trajectory outperforms spending it on a second trajectory that an external reranker has to choose between, because the external reranker is trained only on fully visible images and provides a noisier selection signal than the model's own jointly-trained text encoder.

\textit{Second, scoring at intermediate timesteps converts that advantage into a throughput win.} External reranking must decode every candidate to completion before scoring, so doubling $K$ roughly doubles compute. SD scores at $t_s=8$ of 64 steps (12.5\% decoded), so the remaining 87.5\% is spent only on the surviving candidate. At $K=9$ this yields FID 4.25 and CS 30.1 at 1.86 img/s, which is a 5.6\% FID improvement and 1.7\% CS improvement with higher throughput than external reranking with $K=2$ (1.69 img/s). Under fixed NFE, increasing $K$ from 1 to 17 changes throughput by less than 5\%.


\subsubsection{Scaling Behavior}
\input{results/scaling}
Fig.~\ref{fig:scaling} shows that DREAM scales effectively for both representation and generation. Linear probing accuracy increases monotonically with model size, following standard scaling trends. Generation quality likewise improves: FID drops from 5.67 (DREAM-B) to 3.89 (DREAM-G), matching purely generative baselines such as FLUID. Semantically Aligned Decoding provides consistent gains across all scales, enabling DREAM to surpass FLUID by large margins (+5.1\% on \textbf{L}, +5.1\% on \textbf{H}, and +6.0\% on \textbf{G}). 


\subsubsection{Comparison with Previous Systems}

We compare DREAM against recent T2I and unified multimodal models in Table~\ref{tab:understanding_generation}, separating CC12M-trained models from those trained on substantially larger datasets. Most prior T2I models report only generation quality. Our FLUID re-implementation on CC12M is illustrative, achieving competitive generation (FID 4.53 on CC12M, 9.62 on MS-COCO) yet poor ImageNet-1K transfer (LP 48.1\%), confirming that T2I objectives alone are insufficient for strong discriminative features. 

Among unified models, 4M-L underperforms DREAM on in-domain CC12M generation (FID 11.9 vs.\ 4.25) while leveraging multimodal pseudo-labels distilled from multiple frozen pretrained networks; REPA improves FID over FLUID by aligning with a pretrained encoder but sacrifices discriminative quality (LP 62.5\% vs.\ 72.7\%). Larger LLM-based unified models trained on substantially more data, Show-o-1.3B (35M images), Janus-1.3B (multi-stage with $\sim$tens of M generation samples), and Transfusion-7B ($\sim$3.5B images), achieve stronger MS-COCO FID (9.24, 8.53, 6.78 respectively) than DREAM (10.4), but none of them report results on the standard discriminative transfer suite (linear probing, few-shot, segmentation, depth). The pattern is clear: existing unified models either sacrifice discriminative quality for generation (REPA, 4M-L) or are evaluated only on generation (Show-o, Janus, Transfusion), leaving the trade-off implicit. DREAM is the only entry in this comparison that reports and improves on both axes simultaneously, achieved with a single-stage training recipe and a CC12M-scale corpus.

%% file: results/masking_schedule.tex
\begin{wraptable}{r}{0.5\textwidth}
  \vspace{-1em}
  \centering
  \small
  \caption{\textbf{Component and Schedule Ablations.} Isolating the contribution of masking warmup, diffusion reconstruction, and masking schedule design. DREAM (WM + reconstruction) achieves the best balance between representation (LP) and generation (FID).}
  \label{tab:ablate_schedule}
  \begin{tabularx}{\linewidth}{l *{2}{Y}}
    \toprule
     & \textbf{LP} $\uparrow$ & \textbf{FID} $\downarrow$ \\
    \midrule
    \multicolumn{3}{l}{\textit{Baselines}} \\
    FLUID                       & 48.1          & 4.53 \\
    CLIP                        & 71.6          & N/A \\
    \midrule
    \multicolumn{3}{l}{\textit{DREAM (varying masking schedule)}} \\
    FX                          & 4.6           & N/A \\
    CD                          & \textbf{73.1} & 5.80 \\
    WM (ours) & 72.5        & \textbf{4.57} \\
    \midrule
    DREAM (no recon.)  & 70.8          & N/A \\
    \bottomrule
  \end{tabularx}
  \vspace{-1em}
\end{wraptable}

%% file: results/classification.tex
\begin{table*}[t]
\centering
\footnotesize
\setlength{\tabcolsep}{3pt}
\renewcommand{\arraystretch}{0.9}
\begin{threeparttable}
\caption{
\textbf{Unified evaluation of Visual Understanding and Text-to-Image generation.}
Comparison of models trained on CC12M across two paradigms: (a) \emph{Visual Understanding}, measured by Top-1 accuracy (\%) on ImageNet and its robustness benchmarks; and (b) \emph{Text-to-Image Generation}, measured by FID and CLIP Score (CS) on CC12M and zero-shot MS-COCO. FID and CS for CC12M were computed over 50K samples, and FID and CS for MS-COCO were computed over 30K samples.
\xmark~indicates models inherently incapable of Text-to-Image generation.
}
\label{tab:unified_baselines}
\vspace{2pt}
\begin{tabularx}{\textwidth}{l c *{9}{Y} | Y | *{4}{Y}}
\toprule
& \multicolumn{11}{c|}{\textbf{Visual Understanding}} & \multicolumn{4}{c}{\textbf{Text-to-Image Generation}} \\
\cmidrule(lr){2-12} \cmidrule(lr){13-16}
& \multicolumn{1}{c}{\textbf{LP $\uparrow$}} & \multicolumn{10}{c|}{\textbf{FT $\uparrow$}} & \multicolumn{2}{c}{\textbf{CC12M}} & \multicolumn{2}{c}{\textbf{MS-COCO}} \\
\cmidrule(lr){2-2} \cmidrule(lr){3-12} \cmidrule(lr){13-14} \cmidrule(lr){15-16}
\textbf{Model} &
\mbox{IN-1K} &
\mbox{IN-1K} &
\mbox{IN-A} &
\mbox{IN-R} &
\mbox{IN-S} &
\mbox{IN-H} &
MF &
T7 &
Top &
ON &
Avg. &
\mbox{FID $\downarrow$} &
\mbox{CS $\uparrow$} &
\mbox{FID $\downarrow$} &
\mbox{CS $\uparrow$} \\
\midrule[1pt]
MAR   & 50.7 & 80.4 & 20.6 & 44.6 & 29.4 & 21.1 & 69.0 & 77.0 & 81.7 & 37.4 & 51.2 & \xmark & \xmark & \xmark & \xmark \\
FLUID & 48.1 & 80.3 & 20.2 & 44.2 & 28.6 & 21.1 & 68.9 & 77.3 & 81.6 & 37.6 & 51.1 & 4.53 & 30.0 & \textbf{9.62} & 30.7 \\
CLIP  & 71.6 & 81.1 & 24.3 & 53.9 & 40.8 & 23.3 & 69.3 & 77.1 & 81.7 & 37.8 & 54.4 & \xmark & \xmark & \xmark & \xmark \\
REPA  & 62.5$^\dagger$ & 81.7 & 27.7 & 51.2 & 37.2 & 23.1 & 71.0 & 79.1 & 83.1 & 39.1 & 54.8 & 4.42 & 29.9 & 10.0 & 30.7 \\
\midrule
\cellcolor{green!10}\textbf{DREAM} &
\multirow{2}{*}{\cellcolor{green!10}\textbf{72.7}} &
\multirow{2}{*}{\cellcolor{green!10}\textbf{82.7}} &
\multirow{2}{*}{\cellcolor{green!10}\textbf{32.8}} &
\multirow{2}{*}{\cellcolor{green!10}\textbf{55.3}} &
\multirow{2}{*}{\cellcolor{green!10}\textbf{42.0}} &
\multirow{2}{*}{\cellcolor{green!10}\textbf{26.0}} &
\multirow{2}{*}{\cellcolor{green!10}\textbf{71.7}} &
\multirow{2}{*}{\cellcolor{green!10}\textbf{79.4}} &
\multirow{2}{*}{\cellcolor{green!10}\textbf{83.5}} &
\multirow{2}{*}{\cellcolor{green!10}\textbf{41.3}} &
\multirow{2}{*}{\cellcolor{green!10}\textbf{57.2}} &
\cellcolor{green!10}4.57 & \cellcolor{green!10}29.1 & \cellcolor{green!10}10.0 & \cellcolor{green!10}30.5 \\[3pt]
\cellcolor{green!10}\textcolor{red}{\textbf{+SD}} &
\cellcolor{green!10} & \cellcolor{green!10} & \cellcolor{green!10} &
\cellcolor{green!10} & \cellcolor{green!10} & \cellcolor{green!10} &
\cellcolor{green!10} & \cellcolor{green!10} & \cellcolor{green!10} &
\cellcolor{green!10} & \cellcolor{green!10} &
\cellcolor{green!10}\textcolor{red}{\textbf{4.25}} & \cellcolor{green!10}\textcolor{red}{\textbf{30.1}} & \cellcolor{green!10}\textcolor{red}{10.4} & \cellcolor{green!10}\textcolor{red}{\textbf{31.5}} \\
\bottomrule
\end{tabularx}
\vspace{2pt}
\raggedright
{\footnotesize
$^\dagger$~REPA's linear probe was additionally evaluated on the CLIP-aligned encoder layer, yielding a slightly higher accuracy of 64.4.
}
\end{threeparttable}
\vspace{-3mm}
\end{table*}

%% file: results/unified_results.tex
\begin{table}[t]
\centering
\footnotesize
\setlength{\tabcolsep}{3pt}
\newcolumntype{Y}{>{\centering\arraybackslash}X}
\newcolumntype{W}{>{\hsize=1.5\hsize\centering\arraybackslash}X}
\newcolumntype{N}{>{\hsize=0.875\hsize\centering\arraybackslash}X}
\caption{\textbf{System-level comparison of T2I and unified models.}
We evaluate models on (1) discriminative representations measured by linear probing (``LP'') and fine-tuned (``FT'') accuracy on ImageNet-1K, and (2) T2I generation with FID and CLIP Score (CS) on CC12M and zero-shot MS-COCO.
Metrics shown in \textcolor{blue}{blue} are computed over 40K samples.
``\xmark'' indicates models without text-to-image capability, and ``--'' denotes unreported results.
DREAM maintains competitive representation and generation quality against methods that use additional supervision$^\dagger$.}
\label{tab:understanding_generation}
\begin{tabularx}{\textwidth}{@{} >{\centering\arraybackslash}p{2.0cm} l N N W N N N @{}}
\toprule
 & \multirow{3}{*}{\textbf{Model}} & \multicolumn{2}{c}{\textbf{Discriminative}} & \multicolumn{4}{c}{\textbf{T2I Generation}} \\
\cmidrule(lr){3-4} \cmidrule(lr){5-8}
 &  & \multicolumn{2}{c}{\textbf{IN-1K}} & \multicolumn{2}{c}{\textbf{CC12M-50K}} & \multicolumn{2}{c}{\textbf{MS-COCO-30K}} \\
\cmidrule(lr){3-4} \cmidrule(lr){5-6} \cmidrule(lr){7-8}
 &  & \textbf{LP$\uparrow$} & \textbf{FT$\uparrow$} & \textbf{FID$\downarrow$} & \textbf{CS$\uparrow$} & \textbf{FID$\downarrow$} & \textbf{CS$\uparrow$} \\
\midrule
\multirow{6}{*}{\makecell{T2I}}
 & LPL~\cite{berrada2025boosting}         & --   & --   & 6.22 & 25.1 & --   & --   \\
 & SDXL~\cite{ifriqi2024improved}        & --   & --   & 8.53 & --   & --   & \textcolor{blue}{25.4} \\
 & mmDiT-SD3~\cite{ifriqi2024improved}   & --   & --   & 7.54 & --   & --   & \textcolor{blue}{24.8} \\
 & mmDiT-Imp~\cite{ifriqi2024improved}   & --   & --   & 6.79 & --   & --   & \textcolor{blue}{26.6} \\
 & GALIP~\cite{tao2023galip}       & --   & --   & --   & --   & 14.6 & 29.3 \\
 & SCAD~\cite{kobayashi2025efficiency}        & --   & --   & --   & --   & 12.3 & 27.6 \\
 & FLUID       & 48.1 & 80.3 & 4.53 & 30.0 & 9.62 & 30.7 \\
\midrule
\multirow{7}{*}{\makecell{Unified}}
 & \cellcolor{gray!25}Show-o-1.3B$^\dagger$~\cite{xie2024showo}
   & \cellcolor{gray!25}--
   & \cellcolor{gray!25}--
   & \cellcolor{gray!25}-- & \cellcolor{gray!25}--
   & \cellcolor{gray!25}9.24 & \cellcolor{gray!25}-- \\
 & \cellcolor{gray!25}Janus-1.3B$^\dagger$~\cite{wu2025janus}
   & \cellcolor{gray!25}--
   & \cellcolor{gray!25}--
   & \cellcolor{gray!25}-- & \cellcolor{gray!25}--
   & \cellcolor{gray!25}8.53 & \cellcolor{gray!25}-- \\
 & \cellcolor{gray!25}Transfusion-7B$^\dagger$~\cite{zhou2024transfusion}
   & \cellcolor{gray!25}--
   & \cellcolor{gray!25}--
   & \cellcolor{gray!25}-- & \cellcolor{gray!25}--
   & \cellcolor{gray!25}6.78 & \cellcolor{gray!25}-- \\
 & \cellcolor{gray!25}4M-L$^\ddagger$~\cite{mizrahi20234m}
   & \cellcolor{gray!25}--
   & \cellcolor{gray!25}\textbf{86.6}
   & \cellcolor{gray!25}(11.9$*$) & \cellcolor{gray!25}18.9
   & \cellcolor{gray!25}30.1 & \cellcolor{gray!25}23.1 \\
 & REPA
   & 62.5
   & 81.7
   & 4.42 (5.40$*$) & 29.9
   & 10.0 & 30.7 \\
 & \cellcolor{green!10}\textbf{{DREAM + SD}}
   & \cellcolor{green!10}\textbf{72.7}
   & \cellcolor{green!10}82.7
   & \cellcolor{green!10}\textbf{\mbox{4.25 (5.16$*$)}}
   & \cellcolor{green!10}\textbf{30.1}
   & \cellcolor{green!10}\textbf{10.4}
   & \cellcolor{green!10}\textbf{31.5} \\
\bottomrule
\end{tabularx}
\raggedright
{\small
$^\dagger$ Trained with extra labels or additional supervision (e.g., pretrained LLMs, larger corpora). \\
$^\ddagger$ Trained on CC12M with multimodal pseudo-labels distilled from frozen pretrained networks. \\
$*$ FID computed over 30K samples.
}
\vspace{-1em}
\end{table}

%% file: results/efficiency.tex
\begin{wraptable}{r}{0.5\textwidth}
\vspace{-10mm}
\centering
\footnotesize
\renewcommand{\arraystretch}{1.1}
\setlength{\tabcolsep}{3pt}
\caption{\textbf{Efficiency of DREAM decoding under fixed compute budgets (A100 GPU).} We compare the throughput (images/sec), FID, and CLIP Score (CS) for varying candidate counts~$K$ against using pretrained CLIP as external reranker. \textbf{SD}: Semantically Aligned Decoding.}
\label{tab:dream_k}
\begin{tabular}{@{}p{0.30\textwidth}ccc@{}}
\toprule
Method & Thr.\ $\uparrow$ & FID $\downarrow$ & CLIP $\uparrow$ \\
\midrule
\multicolumn{4}{@{}l}{\textit{64 NFE budget}} \\
\quad DREAM (K=1, T=64) & 3.39 & 4.57 & 29.1 \\
\midrule
\multicolumn{4}{@{}l}{\textit{128 NFE budget}} \\
\quad DREAM (K=1, T=128) & 1.77 & 4.34 & 29.4 \\
\rowcolor{gray!25}\quad DREAM (K=2) + Ext.\ CLIP$^\dagger$ & 1.69 & 4.50 & 29.6 \\
\rowcolor{green!10}\quad DREAM (K=9) + SD & 1.86 & \textbf{4.25} & 30.1 \\
\quad DREAM (K=17) + SD & 1.85 & 4.37 & \textbf{30.3} \\
\bottomrule
\end{tabular}
\parbox{\linewidth}{\small $^\dagger$ Uses an external model to select image candidates.\\
\phantom{$^\dagger$ }All $K>1$ rows use $T=64$.}
\vspace{-6mm}
\end{wraptable}

%% file: results/scaling.tex
\begin{wrapfigure}{r}{0.5\textwidth}
  \vspace{-6mm}
  \centering
  \includegraphics[width=\linewidth]{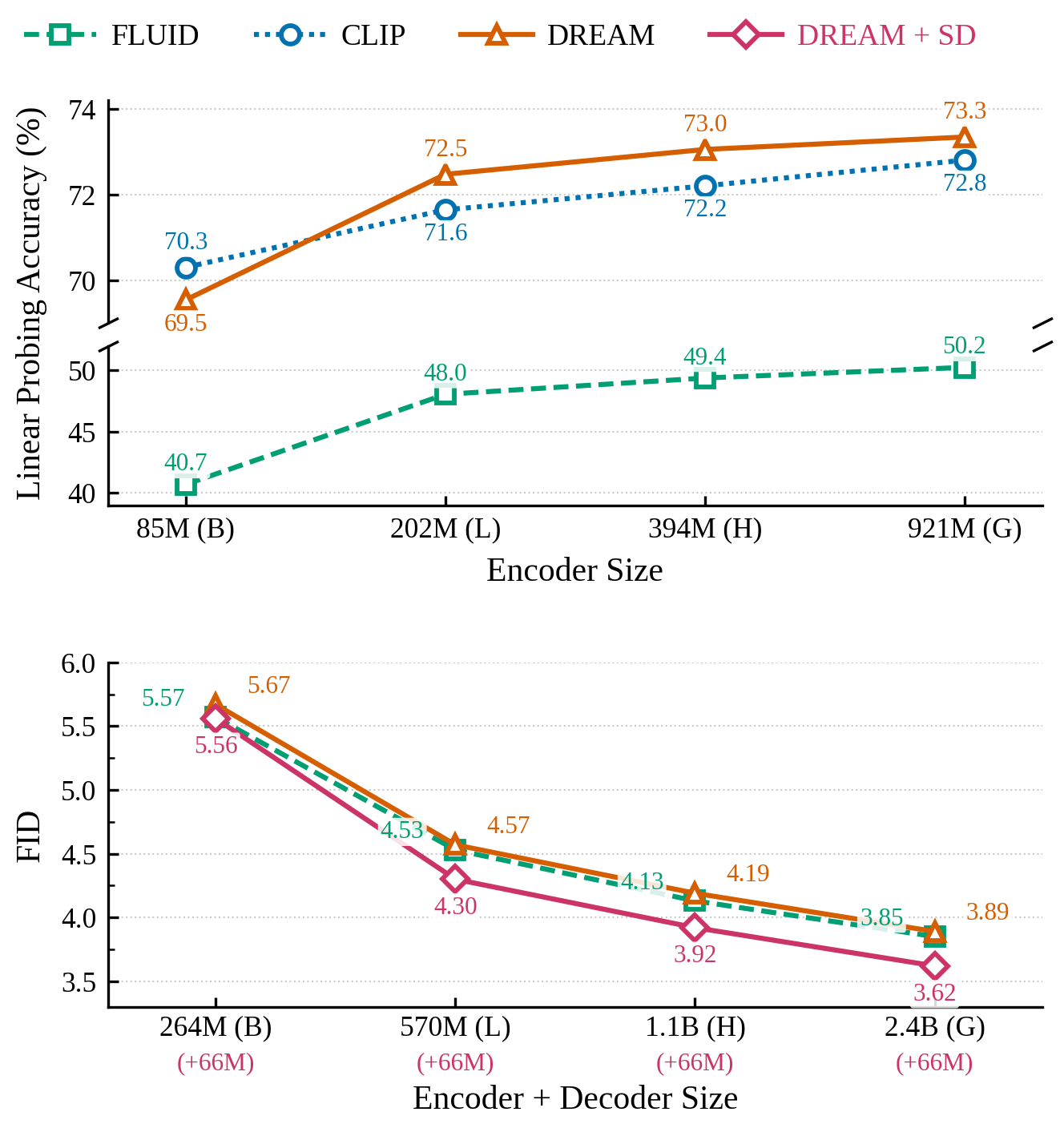}
  \caption{
    \textbf{Performance of \textsc{DREAM} across different model sizes (B/L/H/G) for $\sigma{=}0.45$.}
    \textbf{Top:} Linear Probing on IN-1K.
    \textbf{Bottom:} FID on CC12M.}
  \label{fig:scaling}
  \vspace{-6mm}
\end{wrapfigure}

%% file: sections/discussion.tex
\section{Conclusion}
\label{sec:conclusion}
We introduced DREAM, a unified multimodal framework for visual understanding and text-to-image generation. By reconciling text-image contrastive and generative objectives through Masking Warmup during training and exploiting the resulting joint representations via Semantically Aligned Decoding at inference, we show that the two objectives are synergistic within a single end-to-end trainable architecture. We hope our work will motivate the research community to pursue end-to-end unification of representation and generation across other modalities.


%% file: sections/appendix.tex
\section{Discussion}

\subsection{Limitations}
\label{sec:limitations}
\paragraph{Data scale.}
All experiments use CC12M ($\sim$11.3M pairs), a controlled setting chosen to isolate the effect of single-stage joint optimization with an unfrozen encoder. Comparisons at the hundreds-of-millions scale used by SOTA unified models (e.g., Janus, Harmon) would conflate training objective with data scale. Fig.~\ref{fig:scaling} shows gains hold consistently from $\sim$264M to $\sim$2.4B parameters, suggesting DREAM is well-positioned to benefit from larger data. Empirical verification at that scale remains future work.

\paragraph{Computational overhead.}
DREAM-L incurs $\sim$4\% additional training time and GPU memory over FLUID-L, and contributes an 11.6\% parameter increase relative to FLUID-L. While this overhead is modest given the joint discriminative–generative gains, scaling down the CLIP text encoder is a natural direction for future work. 

\subsection{Broader Impacts}
\label{sec:broader_impacts}
\paragraph{Positive Societal Impacts.}
By enabling end-to-end joint training without frozen encoders, DREAM reduces the engineering and compute overhead typically required to obtain both a strong vision encoder and a competitive text-to-image generator. The training-time cost over a generation-only baseline is modest ($\sim$4\% additional time and memory), which lowers the barrier for academic and resource-constrained groups to study unified multimodal models. Improved visual representations---reflected in gains on classification, semantic segmentation, and depth estimation---can also support beneficial downstream applications such as accessibility tools, scientific image analysis, and assistive technologies.

\paragraph{Negative Societal Impacts.}
DREAM produces a high-quality text-to-image generator and therefore inherits risks shared by this class of models: generation of non-consensual or misleading imagery for disinformation, and amplification of biases present in web-scale image--text training data, which can manifest as stereotyped or uneven depictions across demographic groups. Failure modes such as hallucinated objects or attribute-binding errors may also lead users to over-trust incorrect outputs in sensitive settings (e.g., journalism, education).

\clearpage
\section{Additional Results}

\subsection{Qualitative Results}

\begin{figure}[h]
  \centering
  \includegraphics[width=\linewidth]{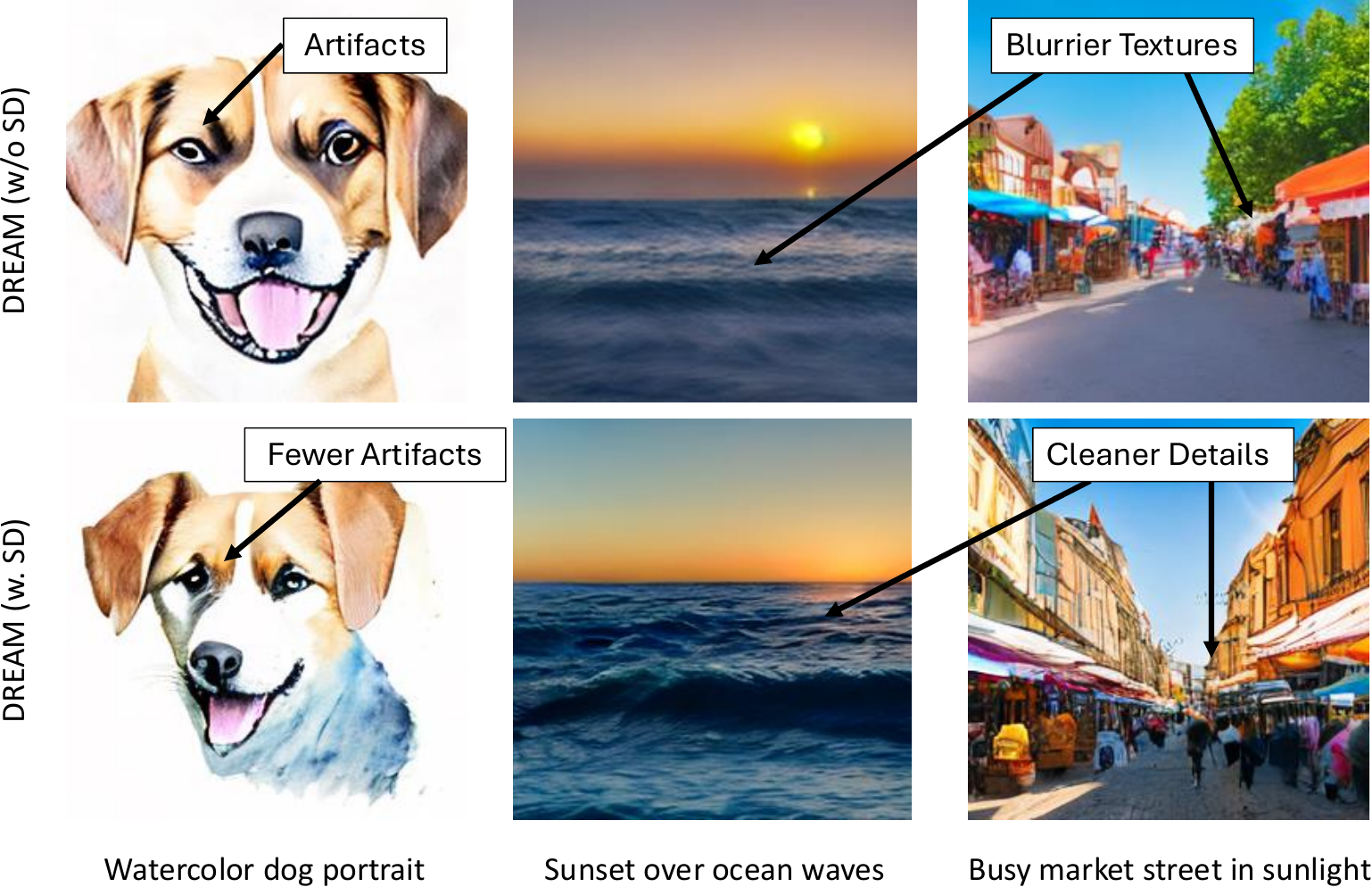}
    \caption{\textbf{Examples of images generated by DREAM with and without Semantically Aligned Decoding (SD).} 
    Without Semantically Aligned Decoding, the outputs exhibit less coherent structure and more low-level blur. 
    Applying Semantically Aligned Decoding produces images with clearer details and improved consistency with the prompt, in line with the gains observed in FID and CLIP scores.}
  \label{fig:semantic_decoding_visual}
    \vskip -0.2em
\end{figure}

We visualize the generated images to understand the effect of Semantically Aligned Decoding. 
Fig.~\ref{fig:semantic_decoding_visual} shows that with Semantically Aligned Decoding, the generated images tend to exhibit more stable structure and fewer low-level artifacts, consistent with the improvements observed in FID.

This effect aligns with the role of the text-guided retrieval step: since intermediate latents are trained to be semantically aligned with the text encoder, retrieving candidates conditioned on the prompt encourages selecting latents that both match the intended content and reconstruct cleanly. Our semantic segmentation results further indicate that the vision encoder encodes strong object-level cues, suggesting that Semantically Aligned Decoding primarily reinforces this semantic grounding during inference.

\begin{figure*}[ht]
  \centering
  \includegraphics[width=0.85\linewidth]{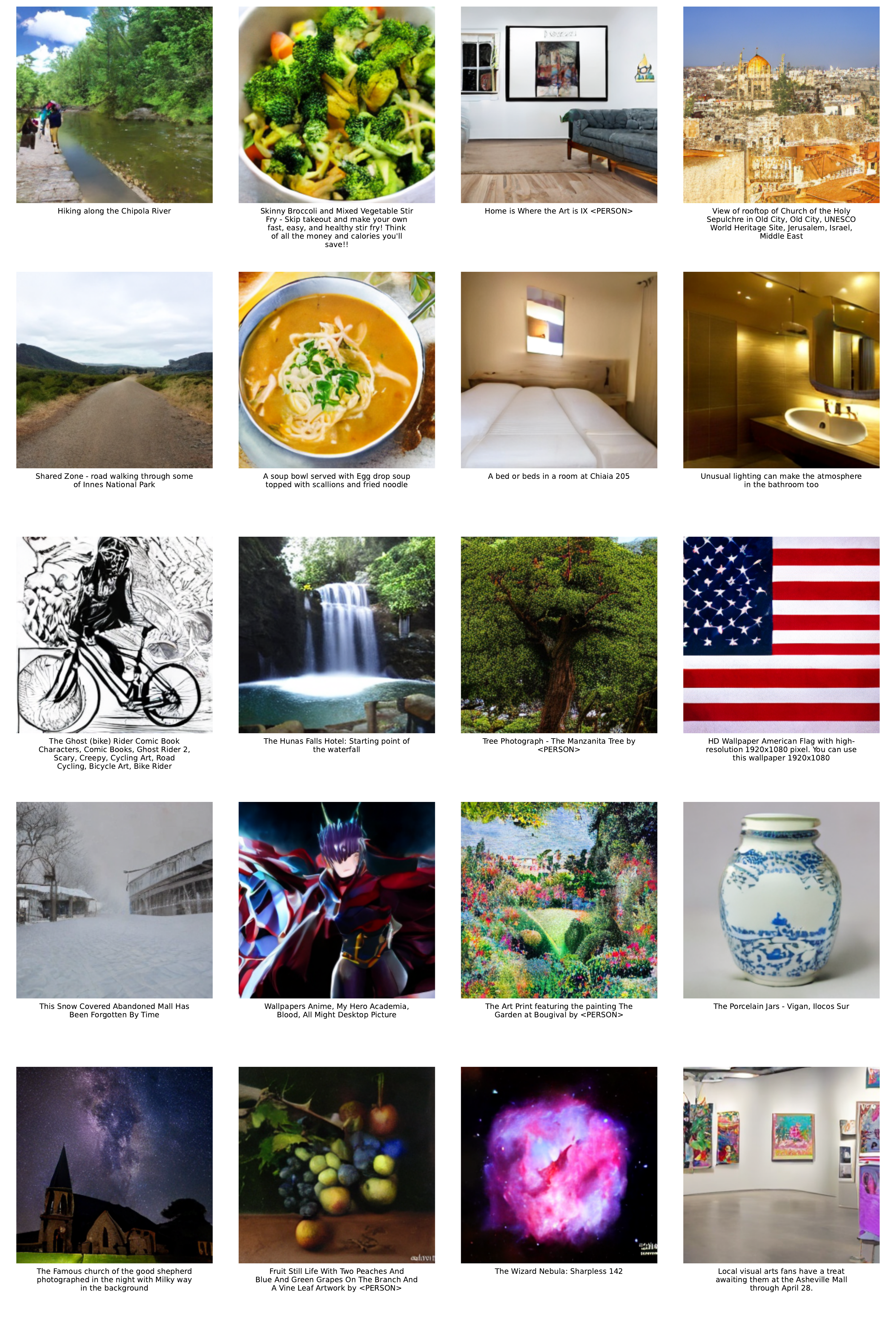}
  \caption{Images generated by DREAM-L (0.57B) model from CC12M captions with CFG = 5.0.}
  \label{fig:gen_img_l}
\end{figure*}

\begin{figure*}[ht]
  \centering
  \includegraphics[width=0.85\linewidth]{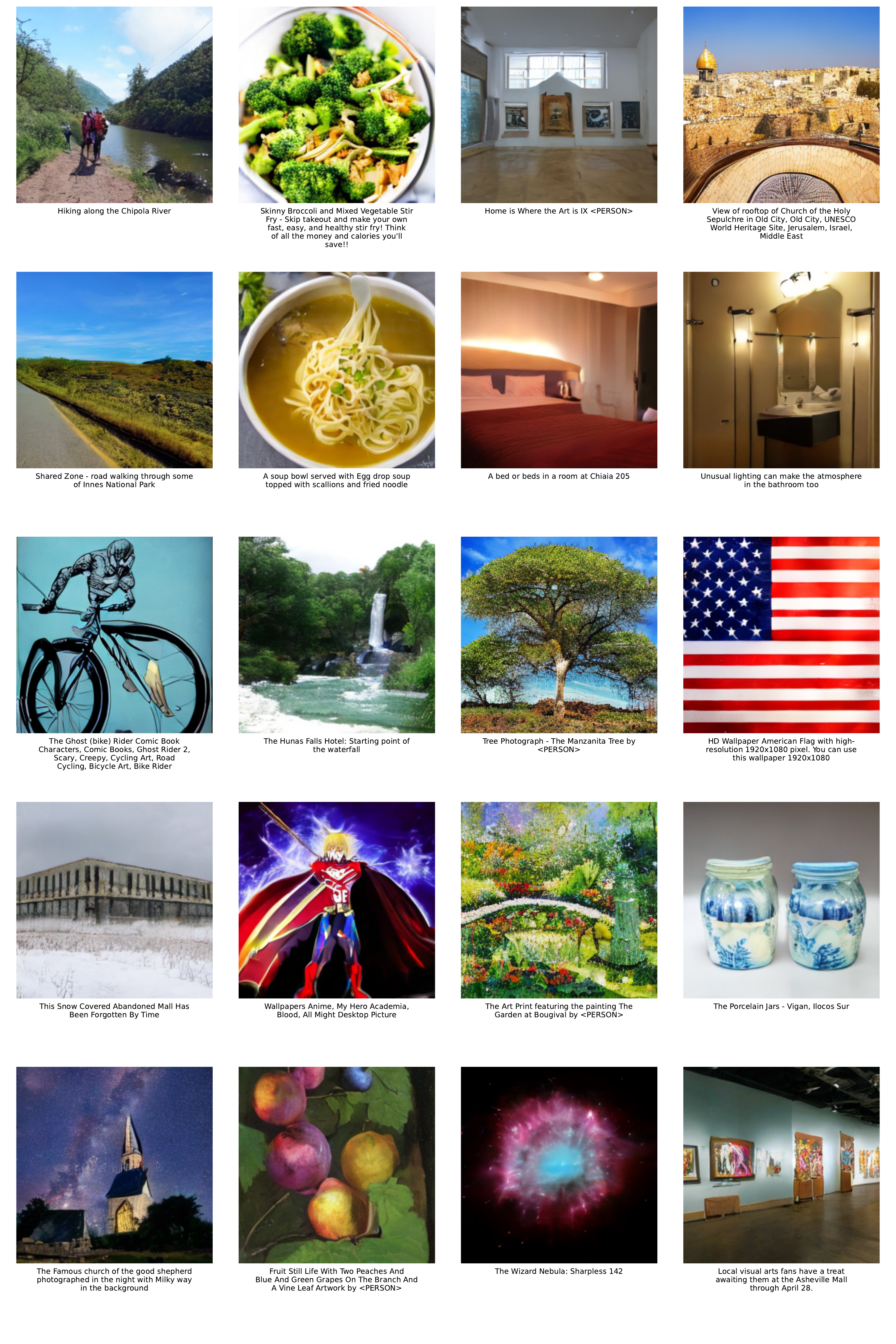}
  \caption{Images generated by DREAM-H (1.1B) model from CC12M captions with CFG = 5.0.}
  \label{fig:gen_img_h}
\end{figure*}

\begin{figure*}[h]
  \centering
  \includegraphics[width=0.85\linewidth]{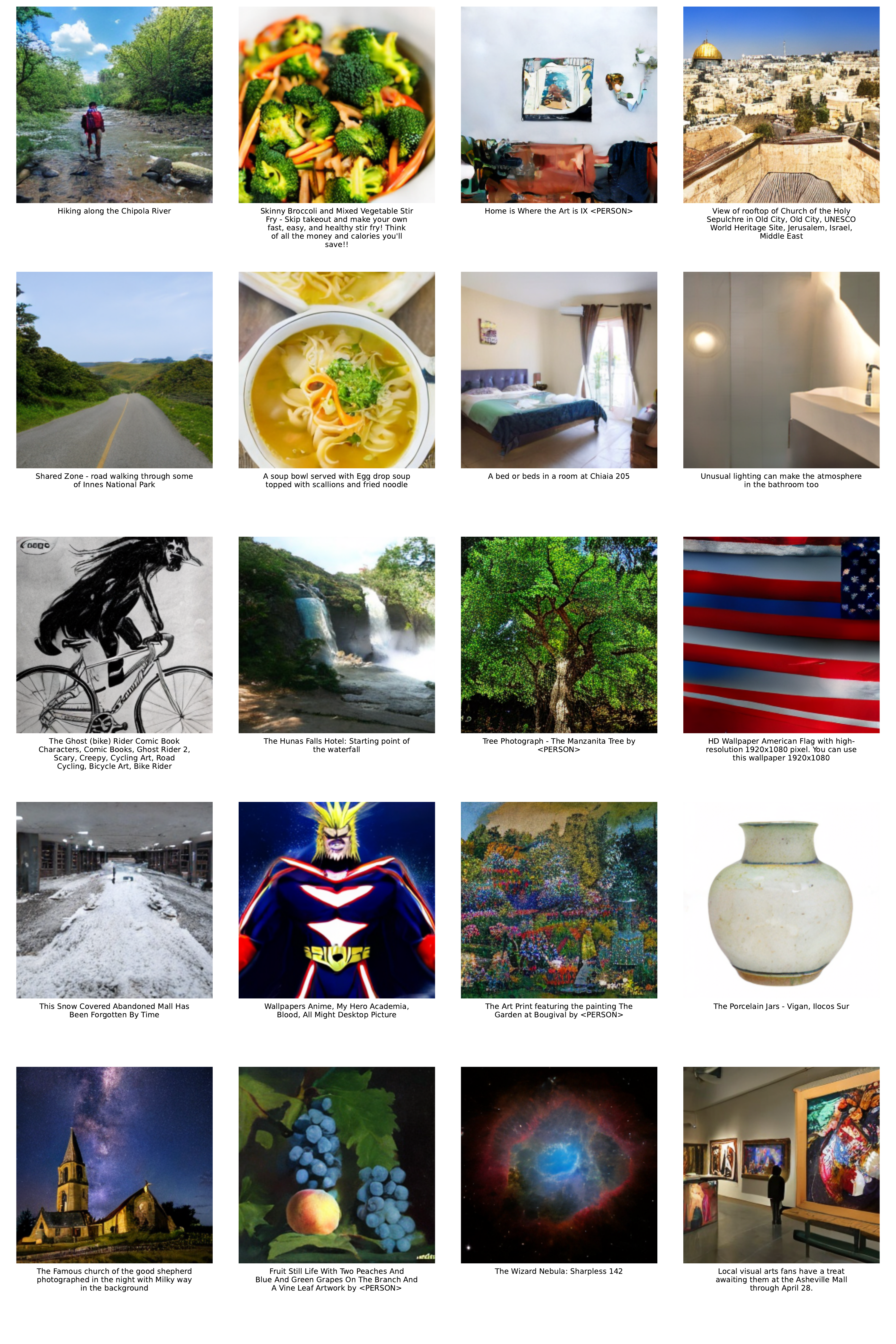}
  \caption{Images generated by DREAM-G (2.4B) model from CC12M captions with CFG = 5.0.}
  \label{fig:gen_img_g}
\end{figure*}

As a result, the decoding process naturally favors samples that are semantically coherent with the prompt while maintaining stronger distribution-level realism.
Additional qualitative examples across different model scales are included in Fig.~\ref{fig:gen_img_l}, \ref{fig:gen_img_h}, and \ref{fig:gen_img_g}.

\subsection{Quantitative Results}

\subsubsection{Stability of Masking Warm-up}
\label{sec:masking_stability}

\begin{figure}[h]
  \centering
  \includegraphics[width=\linewidth]{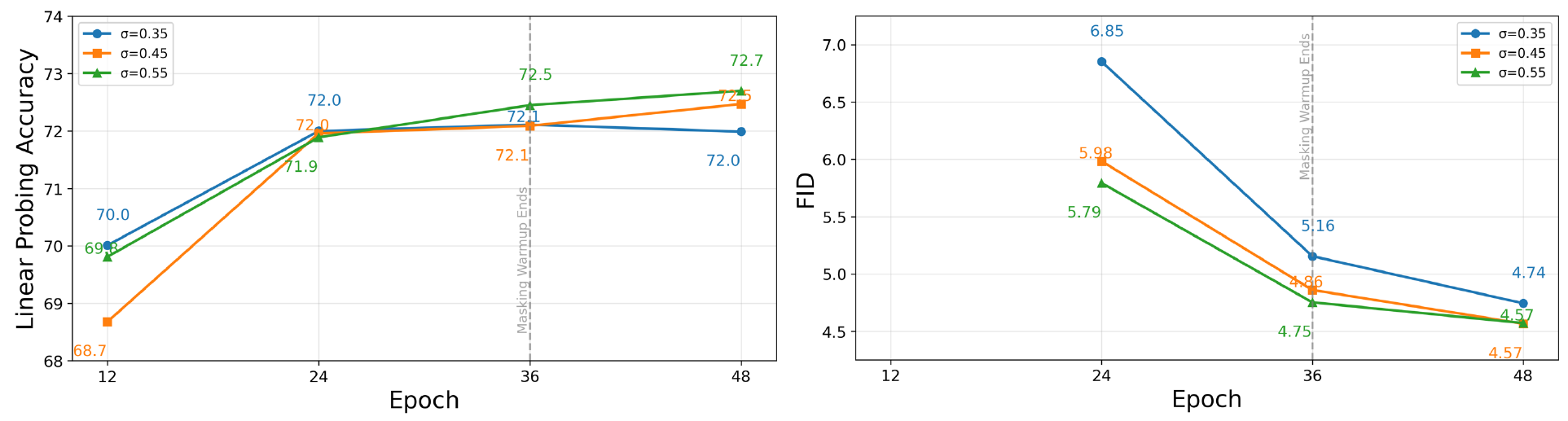}
  \caption{
    \textbf{Performance of DREAM across different masking standard deviations.}
    \textbf{Top:} Linear Probing on IN-1K.
    \textbf{Bottom:} FID on CC12M-50K without Semantically Aligned Decoding. We do not plot the FID values for epoch 12 since the training has not converged and the FID values are too high (FID $>$ 50). The curves show that when $\sigma > 0.35$, both Linear Probing and FID performance are stable with increasing training duration even after masking warm-up ends.}
  \label{fig:stability}
    \vskip -0.2em
\end{figure}

In Fig.~\ref{fig:stability}, we plot Linear Probing accuracy and FID (on CC12M) across training for three masking standard deviations ($\sigma \in {0.35, 0.45, 0.55}$). For $\sigma \in {0.45, 0.55}$, both metrics improve monotonically throughout training. In contrast, with $\sigma = 0.35$, Linear Probing begins to degrade once masking warm-up ends. This suggests that stable unification requires the masking distribution to exceed a certain variance threshold. A larger standard deviation retains more lightly masked samples after masking warm-up ends, ensuring that enough unmasked visual context is available to compute the CLIP loss and maintain discriminative performance.

Overall, these results indicate that given a well-defined masking standard deviation, masking warm-up enables stable joint optimization of discriminative and generative objectives.

\subsubsection{Two-Stage Baselines}
\label{two_stage}

To isolate the effect of staged initialization from the effect of masking warmup, we ablate two-stage pipelines along two axes: the CLIP pretraining duration (epochs 12, 24, and 36), and whether the second stage applies joint fine-tuning (CLIP contrastive loss and MAR diffusion loss together) or generation-only training (MAR diffusion loss alone). We used the same training setting defined in Section~\ref{sec:implementation}.

For the generation-only second stage, we follow FLUID's masking design. For the joint fine-tuning second stage, we additionally introduce the CLIP loss with the same weight used in DREAM's configuration.

\begin{table}[h]
\centering
\caption{Two-stage training results. FID reported as N/A where FID $>$ 50 (generation
collapse). Stage 2 is either generation-only (MAR diffusion loss, no contrastive loss)
or joint training (MAR diffusion loss + CLIP contrastive loss).}
\label{tab:two-stage}
\begin{tabular}{lcccc}
  \toprule
  Stage 2 Training & S1 Epochs & S2 Epochs & LP $\uparrow$ & FID $\downarrow$ \\
  \midrule
  \multirow{3}{*}{Generation-Only} & 12 & 37 & 1.2 & N/A \\
                                   & 24 & 25 & 2.0 & N/A \\
                                   & 36 & 13 & 8.5 & N/A \\
  \midrule
  \multirow{3}{*}{Joint}           & 12 & 37 & 1.8 & N/A \\
                                   & 24 & 25 & 2.5 & N/A \\
                                   & 36 & 13 & 9.1 & N/A \\
  \bottomrule
\end{tabular}
\end{table}

As shown in Table~\ref{tab:two-stage}, varying the Stage 1 duration from 12 to 36 epochs changes the vision encoder's initialization but does not change the outcome. The joint training rows are particularly insightful: even when the CLIP contrastive loss is reintroduced during Stage 2 joint training, both LP and FID collapse across all three initialization strengths. This eliminates the possibility that the failure in the generation-only rows is simply a consequence of missing contrastive regularization in Stage 2.

The failure persists because the root cause lies in the masking distribution design. In the generation-only rows, the high-masking objective overwrites CLIP representations because there is no contrastive signal to anchor them. In the joint training rows, the contrastive signal is present but cannot function effectively under fixed high masking, producing the same collapse. A fixed high-masking regime is structurally incompatible with effective contrastive alignment regardless of initialization strength---consistent with the FX ablation in Table~\ref{tab:ablate_schedule}.

\subsubsection{Few-Shot Performance}
\input{results/fewshot_results}

Following the standard 5-way, 5-shot protocol~\cite{tian2023stablerep, wang2019simpleshot, el2023learning}, we evaluate DREAM across 14 benchmark datasets compared to the other single-stage baselines. DREAM outperforms the other models in most of the datasets, and achieves +4.1\% on average across the 14 datasets as compared to CLIP and provide the results in Table~\ref{tab:fewshot}.

\subsubsection{Dense Prediction}
\label{dense_results}
\input{results/dense_results}

In Table~\ref{tab:segmentation_depth_combined}, we provide the results of Semantic Segmentation on ADE20K~\cite{zhou2017scene} and Depth Estimation on NYU Depth V2~\cite{silberman2012indoor} using linear probes from DINO v3 \cite{simeoni2025dinov3}.

\subsubsection{Layer to apply REPA loss}

\input{appendix/repa_ablation}

Following the method described in \cite{yu2025repa}, we ablated the layer at which the representations are aligned with a pretrained CLIP ViT-L/16 model, and selected the layer with the lowest FID score. We apply a linear probe on the outputs of the vision encoder. As shown in Table~\ref{tab:repa_layer_ablation}, applying REPA at the 6\textsuperscript{th} layer yields the best generative performance, achieving the lowest FID. Interestingly, we observe a consistent trend where shallow alignment layers degrade Linear Probing accuracy.

\subsubsection{CLIP vs.\ DINO Supervision for REPA}
\label{sec:repa_teacher}

Additionally, we compare REPA trained with pretrained CLIP and DINOv2 visual encoders (the default setting in REPA). For fairness, when using DINOv2 supervision, we also ablate and select the optimal feature layer, following the same protocol used for CLIP. We found the optimal layer for REPA loss to be 8th layer.

\input{appendix/repa_backbone}

As shown in Table~\ref{tab:repa_backbone}, REPA supervised with DINOv2 achieves higher linear probing accuracy, indicating stronger semantic representations for downstream recognition tasks. However, this improvement does not translate to better text-to-image generation quality: REPA trained with DINOv2 exhibits worse FID compared to CLIP-supervised REPA. This suggests that while DINOv2 provides stronger category-level representations, CLIP’s text-aligned supervision is more effective for guiding generative models toward high-fidelity, text-consistent image generation.

\subsubsection{Zero-shot Robustness.}
\label{sec:zeroshot_results}

Since DREAM is trained with variable masking ratios, it exhibits improved resilience in highly masked settings. We compare three models—(1) standard CLIP trained without masking, (2) CLIP trained with our masking warm-up schedule (CLIP-M), and (3) DREAM trained with variable-ratio masking—by plotting zero-shot accuracy across varying masking levels.

\input{results/pareto_results}

As shown in Fig.~\ref{fig:zeroshot_pareto}, progressive masking warm-up improves robustness for both CLIP-M and DREAM over CLIP, particularly at moderate masking levels. When more than 20\% of the image is masked, both CLIP-M and DREAM exceed the zero-shot accuracy of CLIP by roughly 0.7\%. Across all masking levels, DREAM further surpasses CLIP-M, demonstrating the additional benefit of its locally grounded representations learned through the reconstruction-based diffusion objective. At extreme occlusion (masking ratio $>0.8$), DREAM’s advantage becomes pronounced: it achieves over 6.2× the zero-shot accuracy of CLIP, highlighting the strong resilience conferred by joint contrastive–generative training.

\subsubsection{Number of Forward Encoder Passes}
\label{sec:nfe_equation}

Semantically Aligned Decoding operates under a fixed compute budget by trading off the number of candidates $K$ against the step at which they are scored. Let $T$ denote the total number of decoding steps (we use $T = 64$ throughout, as described in Section~\ref{sec:inference}), and let $t_s < T$ denote the step at which scoring is performed. 

From step $0$ to $t_s$, all $K$ candidates are decoded in parallel; after scoring, only the winning candidate continues for the remaining $T - t_s$ steps. The total number of forward-encoder (NFE) passes is therefore

\begin{equation}
    \text{NFE} \;=\; \underbrace{K \cdot t_s}_{\text{parallel candidates}} \;+\; \underbrace{(T - t_s)}_{\text{single-candidate decoding}}.
\end{equation}

The fraction of the image decoded at scoring time is $t_s / T$. For the configuration analyzed in Section~\ref{sec:inference_efficiency} ($\text{NFE} = 128$, $K = 9$, $T = 64$), this yields $t_s = 8$, meaning candidate selection occurs with only $8/64 = 12.5\%$ of the image decoded.
 
\vspace{1em}

\subsubsection{Generation Diversity of Semantically Aligned Decoding}
\label{sec:diversity}

To verify that Semantically Aligned Decoding (SD) does not collapse generation toward preferred modes, we conducted a diversity analysis. We generated 20 candidates per prompt for 50 prompts and computed average pairwise cosine dissimilarity within each prompt group using an external CLIP ViT-B/32 encoder. Results are reported in Table~\ref{tab:diversity}.

\begin{table}[h]
  \caption{Intra-prompt generation diversity measured by average pairwise cosine
  dissimilarity using an external CLIP ViT-B/32 encoder. Higher values indicate
  greater diversity.}
  \label{tab:diversity}
  \centering
  \begin{tabular}{lc}
    \toprule
    Condition & Avg.\ Intra-Prompt Diversity \\
    \midrule
    DREAM, $K{=}1$ (no SD)   & 0.122 \\
    External CLIP reranker    & 0.117 \\
    DREAM (SD), $K{=}9$       & 0.119 \\
    \bottomrule
  \end{tabular}
\end{table}

SD and external CLIP reranking produce nearly identical diversity scores (0.119 vs.\ 0.117), both within 0.005 of the no-selection baseline, confirming that SD does not collapse generation toward preferred modes any more than external reranking does. Diversity is preserved because the contrastive alignment space is broad: many visually distinct completions can be semantically consistent with a given prompt, so candidates initialized from independent stochastic seeds naturally remain diverse even after
selection.

\clearpage
\section{Additional Ablations}
We provide further ablations of hyperparameters that enable the unification of representations. Our ablations show that the performance of the model, both in representation learning and text-to-image generation, remains relatively stable across different values, demonstrating that our masking warm-up framework is stable. 

We follow the same default setting as our Analysis Section in the main paper: ablations use DREAM-L trained for 49 epochs on CC12M with the standard configuration: masking standard deviation $\sigma = 0.45$ (which we have established earlier leads to stable joint optimization between representation learning and generation), CLIP loss weight $\lambda = 0.005$, and generation without Semantically Aligned Decoding. We highlight the default settings in gray, and bold the best values in each ablation. 



\subsection{Masking Schedule}
We describe in detail the masking schedules evaluated in Table~\ref{tab:ablate_schedule}. Unless otherwise noted, all schedules use a truncated Gaussian distribution with a standard deviation of 0.45, a maximum masking ratio of 0.75 for computing the CLIP loss (i.e., 75\% of tokens masked), and a minimum masking ratio of 0.5 for computing the MAR loss. The global minimum and maximum masking ratios are fixed at 0 and 1.0, respectively.

\begin{itemize}[noitemsep, topsep=0pt]
    \item \textbf{FX}: Fixed truncated Gaussian distribution centered at 1.0.
    \item \textbf{CD}: A Gaussian distribution whose mean starts at 1.0 and linearly decreases to 0.0 over the first 36 epochs, after which it remains fixed at 0.0.
    \item \textbf{WM}: A Gaussian distribution whose mean starts at 0.0 and linearly increases to 1.0 over the first 36 epochs, after which it remains fixed at 1.0.
\end{itemize}


\subsection{Standard Deviation of Masking}
\label{sec:std_mask}

\begin{table}[h]
\centering
\caption{Standard Deviation of Masking Distribution ($\sigma$)}
\label{tab:sigma_vary}
\begin{tabularx}{0.8\linewidth}{l *{4}{Y}}
\toprule
 & 0.35 & \cellcolor{gray!20}0.45 & 0.55 & UNI\\
\midrule
LP $\uparrow$       & 72.0 & \cellcolor{gray!20}72.5 & \textbf{72.7} & 73.1 \\
FID $\downarrow$    & 4.74 & \cellcolor{gray!20}\textbf{4.57} & \textbf{4.57} & 4.88 \\
\bottomrule
\end{tabularx}
\end{table}

Our masking warm-up samples masking ratios from $\mathcal{N}(\mu, \sigma^2)$ clipped to $[0, 1]$. To study the effect of variance, we vary $\sigma$ and additionally include UNI, a schedule that approximates the high-variance limit via a uniform masking distribution.

As shown in Table~\ref{tab:sigma_vary}, increasing $\sigma$ from 0.35 to 0.45–0.55 improves both representation quality and generative performance: Linear Probing increases from 72.0\% to 72.7\%, while FID decreases from 4.74 to 4.57. Pushing variance to the extreme (UNI) further boosts Linear Probing to 73.1\% but degrades FID to 4.88, improving representation at the cost of generation, rather than both.

Overall, moderate variance in the range $\sigma \in [0.45, 0.55]$ yields the best generative quality while remaining close to the optimal representation performance, demonstrating that the synergistic relationship between text-image contrastive alignment with T2I generation during successful joint unification is stable. 

The unified performance analysis in Section~\ref{results} adopts the setting $\sigma = 0.55$, which matches the best FID (4.57) and provides slightly stronger Linear Probing than $\sigma = 0.45$.

\subsection{Duration of Masking Warm-up}

\begin{table}[h]
\centering
\caption{Masking Warm-up Duration}
\label{tab:masking_duration_vary}
\begin{tabularx}{0.8\linewidth}{l *{3}{Y}}
\toprule
 & 30 & \cellcolor{gray!20}36 & 42 \\
\midrule
LP $\uparrow$       & \textbf{72.5} & \cellcolor{gray!20}\textbf{72.5} & 72.4 \\
FID $\downarrow$    & 4.68 & \cellcolor{gray!20}\textbf{4.57} & 4.71 \\
\bottomrule
\end{tabularx}
\end{table}

We ablated the duration of the masking warm-up. As shown in Table~\ref{tab:masking_duration_vary}, Linear Probing accuracy remains largely stable across warm-up durations of 30–42 epochs, showing only minor differences. In contrast, generative quality exhibits a clear optimum: a 36-epoch warm-up achieves the lowest FID (4.57), outperforming both shorter and longer schedules. This suggests that warm-up duration primarily influences generation fidelity rather than discriminative performance, with a moderate length providing the best balance.

\subsection{Minimum Ratio for Diffusion Loss}

\begin{table}[h]
\centering
\caption{Varying Minimum Masking Ratio for Diffusion Loss ($\gamma$)}
\label{tab:gamma_vary}
\begin{tabularx}{0.8\linewidth}{l *{4}{Y}}
\toprule
 & 0 & 0.25 & \cellcolor{gray!20}0.5 & 0.75 \\
\midrule
LP $\uparrow$       & 72.1 & \textbf{72.7} & \cellcolor{gray!20}72.5 & 71.4 \\
FID $\downarrow$    & 4.95 & 4.73 & \cellcolor{gray!20}\textbf{4.57} & 7.25 \\
\bottomrule
\end{tabularx}
\end{table}

We ablated the minimum masking ratio $\gamma$ used when computing the diffusion loss. As discussed in Section~\ref{training_objectives}, generative training benefits from sufficiently high masking ratios to encourage learning the input distribution.
Table~\ref{tab:gamma_vary} shows that setting $\gamma = 0.5$ (e.g., 50\% of the tokens are masked) yields the best generative performance, achieving the lowest FID (4.57) among the tested configurations. In contrast, Linear Probing accuracy decreases as $\gamma$ increases from 0.25 to 0.75, suggesting that very high minimum masking thresholds reduce the number of low-masking samples seen during training and limit the contribution of reconstruction loss to representation learning.
Overall, moderate values of $\gamma$ provide a better balance between discriminative and generative objectives.

\subsection{Maximum Masking Ratio for CLIP Loss}

\begin{table}[h]
\centering
\caption{Varying Maximum Masking Ratio for CLIP loss ($\phi$)}
\label{tab:phi_vary}
\begin{tabularx}{0.8\linewidth}{l *{4}{Y}}
\toprule
 & 0.25 & 0.5 & \cellcolor{gray!20}0.75 & 1.0\\
\midrule
LP $\uparrow$       & 63.4 & 70.0 & \cellcolor{gray!20}\textbf{72.5} & \textbf{72.5} \\
FID $\downarrow$    & 8.65 & 4.81 & \cellcolor{gray!20}\textbf{4.57} & 4.71 \\
\bottomrule
\end{tabularx}
\end{table}

We ablated the maximum masking ratio $\phi$ used when computing the CLIP loss. As shown in Table~\ref{tab:phi_vary}, increasing $\phi$ from 0.25 to 0.75 leads to consistent gains in Linear Probing accuracy. A larger upper bound allows more samples to retain enough visible content to participate meaningfully in the contrastive objective, strengthening the learned representations.

Overall, the minimum masking ratio for diffusion loss and the maximum masking ratio for CLIP loss together control the degree of overlap between samples that participate in both objectives, thereby influencing the balance between generative reconstruction and discriminative alignment.

\subsection{CLIP Loss Weight ($\lambda$)}

\begin{table}[h!]
\centering
\caption{Effect of CLIP Loss Weight ($\lambda$)}
\label{tab:lambda_vary}
\begin{tabularx}{0.8\linewidth}{l *{3}{Y}}
\toprule
 & 0.002 & \cellcolor{gray!20}0.005 & 0.01 \\
\midrule
LP $\uparrow$       & 72.1 & \cellcolor{gray!20}\textbf{72.5} & 72.0 \\
FID $\downarrow$    & 4.76 & \cellcolor{gray!20}\textbf{4.57} & 4.64 \\
\bottomrule
\end{tabularx}
\end{table}

Additionally, we also ablated the weight on CLIP loss $\lambda$, which balances between representation and generation objectives. Table~\ref{tab:lambda_vary} shows that LP and FID do not change drastically with different weights, demonstrating the stability of DREAM's framework. 

\subsection{CLIP Loss on Buffer Tokens}

\begin{table}[h!]
\centering
\caption{
Application of CLIP loss.
}
\label{tab:clip_loss_buffer}
\setlength{\tabcolsep}{6pt}
\renewcommand{\arraystretch}{1.0}
\begin{tabular}{lcc}
\toprule
 & All Tokens & Buffer Tokens \\
\midrule
\textbf{LP $\uparrow$} & \cellcolor{gray!20}\textbf{72.5} & 72.1 \\
\textbf{FID $\downarrow$} & \cellcolor{gray!20}\textbf{4.57} & 4.57 \\
\bottomrule
\end{tabular}
\end{table}

We further investigate whether restricting the CLIP loss to buffer tokens yields any benefit compared to applying it across both buffer and image tokens. As shown in Table~\ref{tab:clip_loss_buffer}, limiting CLIP supervision to buffer tokens slightly reduces Linear Probing accuracy while leaving FID unchanged.

\subsection{Layer to apply CLIP loss}

\begin{table}[h]
\centering
\caption{CLIP loss layer}
\label{tab:layer_vary}
\begin{tabularx}{0.8\linewidth}{l *{4}{Y}}
\toprule
 & 6 & 8 & 10 & \cellcolor{gray!20}12 \\
\midrule
LP $\uparrow$       & 66.6 & 68.9 & 70.7 & \cellcolor{gray!20}\textbf{72.5} \\
FID $\downarrow$    & 4.51 & 4.71 & 4.53 & \cellcolor{gray!20}\textbf{4.57} \\
\bottomrule
\label{tab:clip_loss_layer}
\end{tabularx}
\end{table}

Similar to REPA, we also ablated the layer to apply CLIP loss. We apply a linear probe on the final layer of the frozen vision encoder (layer 12). As shown in Table~\ref{tab:clip_loss_layer}, deeper layers monotonically improve Linear Probing (66.6\%$\to$72.5\%) while FID remains stable (4.51--4.71). The deepest layer (layer 12) achieves both strong discriminative performance and competitive generation quality, indicating that language alignment at the final encoder layer—where the most abstract semantic features reside—successfully produces semantically-grounded representations without interfering with the decoder's generative capacity.

Notably, our framework demonstrates higher Linear Probing performance than REPA at the same depth (compare with layer 6 in Table~\ref{tab:repa_layer_ablation}), since CLIP loss provides stronger text supervision to the visual encoder than REPA.

\clearpage
\section{Implementation Details}
\label{sec:implementation_appendix}

In the following section, we provide detailed descriptions of our training setup and include pseudocode for DREAM, REPA, and Semantically Aligned Decoding. The full implementation will be made publicly available.

MAR~\cite{li2024autoregressive} is a class-conditional encoder–decoder model in which the encoder takes class labels as input.
FLUID~\cite{fan2024Fluid} is a text-to-image model that uses a transformer architecture operating on continuous tokens, which are cross-attended with text tokens.

DREAM draws architectural inspiration from both MAR and FLUID. We adopt the encoder design from MAR but remove the class-conditioning pathway, ensuring that no textual information enters the visual encoder. On the decoder side, we use FLUID’s architecture and introduce text conditioning only at this stage. This prevents shortcuts between visual features and text and ensures that all discriminative and dense-prediction capabilities learned by the encoder originate solely from visual inputs.

We align the encoder’s visual representations with text using the CLIP contrastive loss~\cite{radford2021clip}. Conceptually, DREAM removes text conditioning from the encoder and instead trains the encoder to predict the relevant textual semantics, reintroducing text conditioning only within the decoder.

\subsection{Image Tokenizer and Detokenizer}
We employ the \texttt{kl-f8-ft-EMA} autoencoder from Stable Diffusion~\cite{rombach2022high}, a widely used continuous tokenizer. The image is first encoded into a $32\times32$ latent grid with 4 channels per token. To maintain consistency with the discrete tokenizer, we group each $2\times2$ block of latent tokens into a single token, resulting in a final sequence of 256 tokens, each with 16 channels. The detokenizer reconstructs the image from the predicted continuous representations.

\subsection{ViT Architecture}

\subsubsection{Vision Encoder}
After the tokenizer, the latent sequence length becomes 256. During training, tokens are masked and dropped before being fed into the encoder. We prepend 64 buffer tokens to the unmasked sequence to ensure stability.
More specifically, we use standard ViT architecture \cite{dosovitskiy2020image}, which consists of a stack of Transformer blocks \cite{dosovitskiy2020image}, where each block consists of a multi-head self-attention block and an MLP block. We use two learnable positional embeddings, one added to the input of the encoder and another added to the input of the decoder.

We use features from the encoder output for classification tasks, such as Linear Probing, Few-Shot Transfer Learning, and Fine-tuning. We apply average pooling to the encoder output for linear classification in Linear Probing and Fine-tuning, and we use the corresponding visual embeddings for semantic segmentation and depth estimation. 

\subsubsection{Text Encoders}
We use two text encoders, each serving a distinct role during training. This separation enables direct comparability with prior work and ensures that our contrastive and generative objectives follow established conventions.

For contrastive alignment, captions are tokenized with the OpenAI CLIP tokenizer (77 tokens) and encoded using a CLIP text transformer, following the setup in~\cite{tian2023stablerep}. The resulting features are projected into the vision encoder’s latent space. 

For generation, captions are tokenized with SentencePiece (128 tokens) and encoded using a frozen T5-XXL model~\cite{raffel2020exploring}. A lightweight 6-layer text aligner projects the T5 embeddings into the decoder’s latent space, providing conditioning for the decoder. This text-conditioning module follows the design used in FLUID~\cite{fan2024Fluid}. 

\subsection{Model Sizes}
We evaluate four model scales: \textbf{Base (B)}, \textbf{Large (L)}, \textbf{Huge (H)}, and \textbf{Giant (G)}. In all settings, the encoder and decoder share the same number of transformer blocks (e.g., a 32-block model allocates 16 blocks to the encoder and 16 to the decoder).

\begin{itemize}
    \item \textbf{Base (B):} 24 blocks, hidden size 768
    \item \textbf{Large (L):} 32 blocks, hidden size 1024
    \item \textbf{Huge (H):} 40 blocks, hidden size 1280
    \item \textbf{Giant (G):} 48 blocks, hidden size 1664
\end{itemize}

The \textbf{Huge} configuration corresponds to a $\sim$1B-parameter encoder-decoder model.  
The \textbf{Giant} configuration ($\sim$2B parameters in total) provides an encoder with approximately 1B parameters (921M), enabling us to study scaling behavior in the 1B-encoder regime.

\subsection{Pre-training}
Section~\ref{sec:settings} summarizes the default pre-training configurations used across all experiments. Unless otherwise specified, all models are trained for 49 epochs.

\subsubsection{DREAM training}
DREAM uses a progressive masking warm-up in which the masking ratio follows a shifting truncated Gaussian distribution. The minimum and maximum masking ratios are fixed at 0 and 1.0, respectively. The mean of the distribution increases linearly from 0 to 1.0 over the first 36 epochs and is then fixed at 1.0 for the remaining epochs. This schedule enables a smooth transition from fully visible inputs to the high-masking regime required for strong generative performance.

\subsubsection{REPA training}
REPA~\cite{yu2025repa} was originally proposed for diffusion transformers, where noisy intermediate encoder states are aligned with clean-image features from a pretrained vision encoder.
In our adaptation for masked image modeling, we instead align the unmasked tokens of the vision encoder with the corresponding features computed from the full image using a pretrained CLIP-L encoder. This provides REPA-style feature supervision in a manner that is compatible with our masked-token training setup.

\subsubsection{CLIP training}
Our CLIP baseline is trained without masking for the full 49 epochs.

\subsubsection{FLUID and MAR training.}
FLUID and MAR baselines are trained for 49 epochs following the MAR masking schedule~\cite{li2024autoregressive}: masking ratios are sampled from a truncated Gaussian centered at 1.0 (100\% masking) with a standard deviation of 0.25, bounded to the range [0.7, 1.0]. This setup allows us to directly evaluate the vision encoder and assess the effect of applying text conditioning only in the decoder.

\subsection{Inference}
\subsubsection{Generation}
DREAM adopts the MAR \cite{li2024autoregressive} decoding procedure, which generates images by predicting a randomly selected set of latent tokens at each step. At inference time, we begin from a fully masked latent sequence and iteratively reduce the masking ratio following a cosine schedule. By default, we use 64 steps in this schedule. In each iteration, DREAM first predicts continuous latent values for all currently masked positions. It then applies token-wise temperature sampling by adding temperature-scaled noise to the predicted latent vectors. Finally, instead of using confidence-based or location-based ordering as in MAGE, DREAM follows MAR's fully randomized ordering: a subset of positions is sampled uniformly and re-masked to match the target masking ratio for the next iteration. This process progressively refines the latent representation until all positions are filled.

\subsubsection{Classifier-free guidance (CFG)}
We adopt classifier-free guidance by randomly removing the text condition for 10\% of training samples, replacing the prompt with a null token. This trains the model to produce both conditional and unconditional predictions using the same diffusion head.

During inference, we run the model twice: once with the null prompt and once with the given text. This produces the unconditional and conditional representations $z_u$ and $z_c$. The guided noise prediction is computed as
\[
\varepsilon = \varepsilon_\theta(x_t \mid t, z_u) \;+\; \omega \cdot \big(\varepsilon_\theta(x_t \mid t, z_c) - \varepsilon_\theta(x_t \mid t, z_u)\big),
\]
where $\omega$ denotes the guidance strength. Since we use a fixed sampling temperature of~1, no additional temperature scaling is applied.

For sampling, we sweep over two CFG schedules: a constant scale and a linearly increasing scale, and report the best CLIP and FID scores. 

\subsubsection{Semantic Segmentation}
For semantic segmentation, we follow the DINOv3~\cite{simeoni2025dinov3} setup. A linear layer is trained on top of the frozen patch features from the vision encoder and evaluated on ADE20K~\cite{zhou2017scene} (150 categories). We report mean IoU and Pixel Accuracy. The linear head is trained for 40k iterations with a learning rate of $1\times10^{-4}$ using a cosine decay schedule with 1.5k warm-up steps.

\subsubsection{Depth Estimation}
For depth estimation, we follow the DINOv2~\cite{oquab2023dinov2} protocol. A linear head is trained over frozen patch features and evaluated on NYU Depth v2~\cite{silberman2012indoor}. We report RMSE and ARel. Training uses 38.4k iterations with a learning rate of $3\times10^{-4}$, a cosine scheduler, and 12.8k warm-up steps.

\subsection{Training Efficiency}
\label{sec:training_efficiency_appendix}

\begin{table}[h]
  \small
  \caption{Training efficiency comparison of large-scale models. Training overhead and peak memory are measured on the same hardware setup.}
  \label{tab:efficiency}
  \centering
  \begin{tabular}{lcccc}
    \toprule
    Model & Total Params & Train GFLOPs/sample & Training Time & Peak GPU Mem \\
    \midrule
    FLUID-L & 570M & 772.8 & $\sim$48 hours & 43,827 MB \\
    REPA-L  & 578M & 825.2 & $\sim$48 hours & 44,872 MB (+2.4\%) \\
    DREAM-L & 636M (+11.6\%) & 877.2 (+13.5\%) & $\sim$50 hours (+4.2\%) & 45,688 MB (+4.2\%) \\
    \bottomrule
  \end{tabular}
\end{table}

Here, we compare the training efficiency between the T2I models. All models were trained on NVIDIA A100 GPUs. The results show that despite incorporating two text encoders, DREAM-L incurs only modest overhead compared to its baselines: a 11.6\% increase in parameters and 13.5\% increase in training compute over FLUID-L, while requiring only 4.2\% more training time and 4.2\% additional GPU memory. Additionally, we note that the increase in total parameters arises from the CLIP text encoder, whose size is fixed across model scales (B/L/H/G), so its relative parameter overhead diminishes as the backbone grows.

These numbers confirm that the additional architectural components of DREAM do not impose a significant practical burden during training

\clearpage
\section{Assets and Licenses}
\input{asset_licenses}

\clearpage
\section{Training Settings}
\label{sec:settings}

\input{appendix/setting/pretraining_clip}
\input{appendix/setting/pretraining_mar_fluid}
\input{appendix/setting/pretraining_repa}
\input{appendix/setting/pretraining_dream}
\input{appendix/setting/linear_probing}
\input{appendix/setting/finetuning}

\clearpage

\section{Pseudocode}
\label{sec:pseudocode}
\input{appendix/pseudocode/dream}

\input{appendix/pseudocode/repa}

\input{appendix/pseudocode/semantically_aligned_decoding}

\clearpage

\clearpage

%% file: results/fewshot_results.tex
\begin{table*}[h!]
\centering
\small
\setlength{\tabcolsep}{4pt}
\renewcommand{\arraystretch}{0.9}
\begin{threeparttable}
\caption{\textbf{Few-shot transfer evaluation of different models}. We report 5-way, 5-shot classification accuracy across 14 datasets. We highlight the best performance of each dataset in bold.}
\label{tab:fewshot}
\vspace{-1mm}
\begin{tabularx}{\textwidth}{l *{14}{Y} | Y}
\toprule
\textbf{Model} &
\rotatebox{90}{DTD \cite{cimpoi14describing}} &
\rotatebox{90}{Caltech-256 \cite{griffin2022caltech}} &
\rotatebox{90}{SUN397 \cite{Xiao:2010}} &
\rotatebox{90}{Food-101 \cite{bossard14}} &
\rotatebox{90}{VOC2007 \cite{everingham2010pascal}} &
\rotatebox{90}{STL-10 \cite{coates2011stl10}} &
\rotatebox{90}{Flowers \cite{Nilsback08}} &
\rotatebox{90}{UC Merced \cite{yang2010bag}} &
\rotatebox{90}{EuroSAT \cite{helber2017eurosat}} &
\rotatebox{90}{Country211 \cite{thomee2016yfcc100m}} &
\rotatebox{90}{RESISC45 \cite{Cheng_2017}} &
\rotatebox{90}{Dogs \cite{KhoslaYaoJayadevaprakashFeiFei_FGVC2011}} &
\rotatebox{90}{MIT Indoors \cite{quattoni2009recognizing}} &
\rotatebox{90}{IN-1K \cite{deng2009imagenet}} &
\rotatebox{90}{\textbf{Average}} \\
\midrule
MAR & 64.9 & 56.1 & 73.0 & 42.3 & 47.2 & 53.1 & 74.6 & 77.3 & 74.9 & 28.9 & 73.8 & 37.7 & 61.2 & 62.3 & 59.1 \\
FLUID & 66.3 & 58.0 & 73.7 & 43.7 & 47.9 & 53.7 & 76.1 & 77.0 & 75.8 & 29.4 & 74.4 & 38.9 & 62.7 & 64.1 & 60.1 \\
CLIP & 81.0 & 95.1 & 95.8 & 86.0 & 79.6 & 96.1 & 96.2 & 92.6 & 82.1 & \textbf{42.9} & 90.4 & 78.2 & 93.7 & 94.3 & 86.0 \\
REPA & 73.2 & 73.6 & 88.1 & 60.2 & 61.2 & 74.6 & 86.5 & 82.4 & 79.2 & 33.8 & 80.4 & 50.2 & 81.2 & 78.9 & 71.7 \\
\midrule
\cellcolor{green!10}\textbf{DREAM} & \cellcolor{green!10}\textbf{88.6} & \cellcolor{green!10}\textbf{96.4} & \cellcolor{green!10}\textbf{97.3} & \cellcolor{green!10}\textbf{86.9} & \cellcolor{green!10}\textbf{95.0} & \cellcolor{green!10}\textbf{99.8} & \cellcolor{green!10}\textbf{95.9} &\cellcolor{green!10}\textbf{98.5} & \cellcolor{green!10}\textbf{95.0} & \cellcolor{green!10}\textbf{42.9} & \cellcolor{green!10}\textbf{93.4} & \cellcolor{green!10}\textbf{81.2} & \cellcolor{green!10}\textbf{94.5} & \cellcolor{green!10}\textbf{96.1} & \cellcolor{green!10}\textbf{90.1} \\
\bottomrule
\end{tabularx}
\end{threeparttable}
\vspace{-2mm}
\end{table*}

%% file: results/dense_results.tex
\begin{table}[h]
    \centering
    \small
    \caption{
    \textbf{Performance of frozen backbones on dense prediction tasks.} Results are reported for semantic segmentation (on ADE20K~\cite{zhou2017scene}) and depth estimation (on NYU Depth V2~\cite{silberman2012indoor}) using linear probes \cite{simeoni2025dinov3} mIoU: mean Intersection-over-Union; PixAcc: pixel accuracy; RMSE: root mean squared error; ARel: absolute relative error.
    }    
    \setlength{\tabcolsep}{3.5pt}
    \begin{tabular}{l|cc|cc}
    \toprule
    & \multicolumn{2}{c|}{\textbf{Sem. Seg. (ADE20K)}} & \multicolumn{2}{c}{\textbf{Depth (NYU Depth V2)}} \\
    \cmidrule(lr){2-3} \cmidrule(lr){4-5}
    \textbf{Model} & mIoU$\uparrow$ & PixAcc$\uparrow$ & RMSE$\downarrow$ & ARel$\downarrow$ \\
    \midrule
    MAR       & 23.4 & 70.6 & 0.75 & 0.25 \\
    FLUID       & 22.1 & 69.5 & 0.76 & 0.26 \\
    CLIP         & 34.9 & 75.3 & 0.64 & 0.20 \\
    REPA         & 32.7 & 75.9 & \textbf{0.60} & \textbf{0.19} \\
    \midrule
    \cellcolor{green!10}\textbf{DREAM} 
                 & \cellcolor{green!10}\textbf{36.8} 
                 & \cellcolor{green!10}\textbf{76.7} 
                 & \cellcolor{green!10}\textbf{0.60} 
                 & \cellcolor{green!10}\textbf{0.19} \\
    \bottomrule
    \end{tabular}
\label{tab:segmentation_depth_combined}
\end{table}

%% file: appendix/repa_ablation.tex
\begin{table}[h]
\centering
\caption{
\textbf{Ablation on the layer to align with pretrained CLIP encoder.}
Applying the loss at earlier (shallower) layers yields better linear probing (LP) performance, 
whereas applying it at later (deeper) layers improves generation quality (lower FID), highlighting the trade-off between representation and generation quality.
}
\label{tab:repa_layer_ablation}
\setlength{\tabcolsep}{6pt}
\renewcommand{\arraystretch}{1.0}
\begin{tabular}{lccc}
\toprule
 & 4 & 6 & 8 \\
\midrule
LP $\uparrow$ & 58.0 & \textbf{62.5} & 66.1 \\
FID $\downarrow$ & 4.71 & \textbf{4.42} & 4.76 \\
\bottomrule
\end{tabular}
\end{table}

%% file: appendix/repa_backbone.tex
\begin{table}[h]
\centering
\caption{Effect of Visual Backbone}
\label{tab:repa_backbone}
\begin{tabular}{lcc}
\toprule
 & \cellcolor{gray!20}CLIP & DINOv2 \\
\midrule
LP $\uparrow$       & \cellcolor{gray!20}62.5 & 68.4 \\
FID $\downarrow$    & \cellcolor{gray!20}4.42 & 4.67 \\
\bottomrule
\end{tabular}
\end{table}


%% file: results/pareto_results.tex
\begin{figure}[h]
\centering
\includegraphics[width=0.7\linewidth]{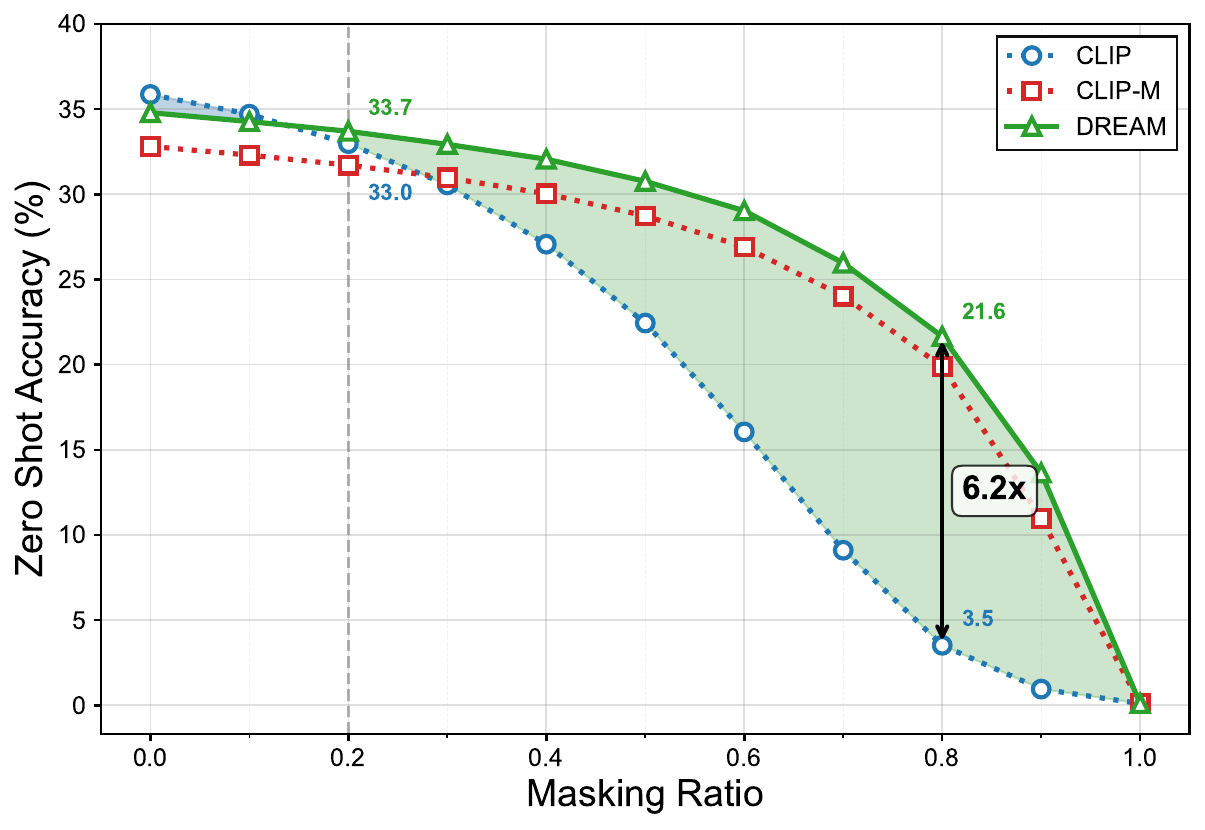}
\caption{\textbf{Comparison of zero-shot accuracy} for CLIP, CLIP with masking warmup (CLIP-M), and DREAM across masking ratios. The green shaded region marks the masking levels at which DREAM surpasses CLIP.}
\label{fig:zeroshot_pareto}
\end{figure}

%% file: asset_licenses.tex
\label{sec:asset_licenses}

This work uses several existing datasets, models, and software libraries which we list below. 

\paragraph{Datasets.}
\begin{itemize}
    \item \textbf{Conceptual 12M (CC12M)}~\citep{changpinyo2021cc12m}. Used for pre-training. URL: \url{https://github.com/google-research-datasets/conceptual-12m}. License: provided by Google LLC ``AS IS'' and freely usable for any purpose with acknowledgement of Google as the data source. CC12M is a list of (image-URL, alt-text) pairs scraped from the web; the dataset distributors do not own the underlying images and we accessed images via the provided URLs subject to the original copyright of each web source.
    \item \textbf{ImageNet-1K (ILSVRC 2012)}~\citep{deng2009imagenet}. Used for linear probing, fine-tuning, and as the source for the few-shot IN-1K split. URL: \url{https://image-net.org/}. License: ImageNet Terms of Access (non-commercial research and educational use only); ImageNet does not own the copyright of the underlying images.
    \item \textbf{ImageNet variants for robustness evaluation:} ImageNet-A~\citep{hendrycks2021nae} (MIT License, \url{https://github.com/hendrycks/natural-adv-examples}), ImageNet-R~\citep{hendrycks2021many} (MIT License, \url{https://github.com/hendrycks/imagenet-r}), ImageNet-Sketch~\citep{wang2019learning} (MIT License, \url{https://github.com/HaohanWang/ImageNet-Sketch}), ImageNet-Hard~\citep{taesiri2023imagenet} (MIT License, \url{https://github.com/taesiri/ImageNet-Hard}), ObjectNet~\citep{barbu2019objectnet} (custom non-commercial research license, \url{https://objectnet.dev/}), and the ImageNetV2 splits MatchedFrequency, Threshold0.7, and TopImages~\citep{recht2019imagenet} (MIT License, \url{https://github.com/modestyachts/ImageNetV2}).
    \item \textbf{MS-COCO}~\citep{lin2014microsoft}. Used for zero-shot text-to-image evaluation (MS-COCO 30K). URL: \url{https://cocodataset.org/}. License: annotations licensed under Creative Commons Attribution 4.0 (CC BY 4.0); the images are not owned by the COCO Consortium and use of the images is subject to the original Flickr Terms of Use.
    \item \textbf{ADE20K}~\citep{zhou2017scene}. Used for semantic segmentation. URL: \url{https://groups.csail.mit.edu/vision/datasets/ADE20K/}. License: image annotations and software released under the BSD-3-Clause License (Copyright 2019 MIT, CSAIL); the underlying images are provided for non-commercial research and educational use, with MIT CSAIL not owning the copyright of the original images.
    \item \textbf{NYU Depth V2}~\citep{silberman2012indoor}. Used for monocular depth estimation. URL: \url{https://cs.nyu.edu/~silberman/datasets/nyu_depth_v2.html}. The original release does not specify a formal license; commonly redistributed preprocessed versions (e.g., FastDepth) are released under the MIT License.
    \item \textbf{Few-shot transfer datasets}:
    DTD~\citep{cimpoi14describing} (research-only, \url{https://www.robots.ox.ac.uk/~vgg/data/dtd/}),
    Caltech-256~\citep{griffin2022caltech} (CC BY 4.0 via CaltechDATA),
    SUN397~\citep{Xiao:2010} (research-only, \url{https://vision.princeton.edu/projects/2010/SUN/}),
    Food-101~\citep{bossard14} (research-only, ETH Zurich),
    PASCAL VOC 2007~\citep{everingham2010pascal} (Flickr Terms of Use; annotations free for research),
    STL-10~\citep{coates2011stl10} (research-only),
    Oxford 102 Flowers~\citep{Nilsback08} (research-only, University of Oxford),
    UC Merced Land Use~\citep{yang2010bag} (public domain; images from USGS),
    EuroSAT~\citep{helber2017eurosat} (MIT License),
    Country211~\citep{thomee2016yfcc100m} (derived from YFCC100M; CC license metadata per image; Country211 split released by OpenAI under MIT License),
    RESISC45~\citep{Cheng_2017} (research-only, \url{https://gcheng-nwpu.github.io/}),
    Stanford Dogs~\citep{KhoslaYaoJayadevaprakashFeiFei_FGVC2011} (research-only, derived from ImageNet),
    MIT Indoor Scenes~\citep{quattoni2009recognizing} (research-only, \url{https://web.mit.edu/torralba/www/indoor.html}).
\end{itemize}

\paragraph{Pretrained models.}
\begin{itemize}
    \item \textbf{Stable Diffusion VAE} (\texttt{kl-f8-ft-EMA})~\citep{rombach2022high}. Used as the continuous image tokenizer. URL: \url{https://github.com/CompVis/stable-diffusion}. License: CreativeML Open RAIL-M.
    \item \textbf{T5-XXL v1.1}~\citep{raffel2020exploring}. Used (frozen) as the decoder-side text encoder for generation. URL: \url{https://huggingface.co/google/t5-v1_1-xxl}. License: Apache License 2.0.
    \item \textbf{OpenAI CLIP}~\citep{radford2021clip}. Used for: (i) the pretrained CLIP-L teacher in the REPA baseline, (ii) the CLIP ViT-B/32 used as an external reference encoder for the diversity analysis (Table~\ref{tab:diversity}), and (iii) the CLIP-style text encoder architecture and tokenizer for our contrastive branch. URL: \url{https://github.com/openai/CLIP}. License: MIT License.
    \item \textbf{DINOv2}~\citep{oquab2023dinov2}. Used as an alternative REPA teacher in the ablation in Section~\ref{sec:repa_teacher}. URL: \url{https://github.com/facebookresearch/dinov2}. License: Apache License 2.0.
\end{itemize}

\paragraph{Code bases and software libraries.}
\begin{itemize}
    \item \textbf{MAR}~\citep{li2024autoregressive}. The encoder architecture, MAR masking strategy, decoding procedure, and diffusion head implementation are adapted from the official MAR codebase. URL: \url{https://github.com/LTH14/mar}. License: MIT License.
    \item \textbf{REPA}~\citep{yu2025repa}. We adapt the representation-alignment objective for the masked-image-modeling setting as a baseline. URL: \url{https://github.com/sihyun-yu/REPA}. License: MIT License.
    \item \textbf{SentencePiece}~\citep{kudo2018sentencepiece}. Used for tokenizing captions for the T5 text encoder. URL: \url{https://github.com/google/sentencepiece}. License: Apache License 2.0.
    \item \textbf{PyTorch} and standard Python scientific libraries (NumPy, SciPy, timm, etc.) under their respective open-source licenses (BSD-3-Clause, BSD, Apache 2.0, etc.).
    \item \textbf{DINOv3}~\citep{simeoni2025dinov3}. The dense-prediction linear-probe protocol of DINOv3 is followed for ADE20K segmentation evaluation. URL: \url{https://github.com/facebookresearch/dinov3}. License: DINOv3 License (custom, non-Apache).
\end{itemize}

%% file: appendix/setting/pretraining_clip.tex

\begin{table}[h]
\centering
\caption{Pre-training settings for CLIP.}
\label{tab:pretrain_clip}
\setlength{\tabcolsep}{12pt} 
\renewcommand{\arraystretch}{1.05}
\begin{tabular}{l l} 
\toprule
\textbf{Config} & \textbf{Value} \\
\midrule
Optimizer & AdamW \\
Base learning rate & $1\times10^{-4}$ \\
Weight decay & 0.02 \\
Momentum & (0.9, 0.95) \\
Batch size & 2048 \\
LR schedule & Constant \\
Warmup epochs & 12 \\
Training epochs & 49 \\
Gradient clip & 3.0 \\
Label dropout & 0.1 \\
Augmentation & Random Horizontal Flipping \\
\bottomrule
\end{tabular}
\end{table}

%% file: appendix/setting/pretraining_mar_fluid.tex

\begin{table}[h]
\centering
\caption{Pre-training settings for MAR, and FLUID.}
\label{tab:pretrain_mar_clip_fluid_full}
\setlength{\tabcolsep}{12pt}
\renewcommand{\arraystretch}{1.05}
\begin{tabular}{l l} 
\toprule
\textbf{Config} & \textbf{Value} \\
\midrule
Optimizer & AdamW \\
Base learning rate & $1\times10^{-4}$ \\
Weight decay & 0.02 \\
Momentum & $(0.9, 0.95)$ \\
Batch size & 2048 \\
LR schedule & Constant \\
Warmup epochs & 12 \\
Training epochs & 49 \\
Gradient clip & 3.0 \\
Label dropout & 0.1 \\
Augmentation & Random Horizontal Flipping \\
\midrule
Masking ratio (min) & 0.7 \\
Masking ratio (max) & 1.0 \\
Masking ratio (std) & 0.25 \\
\bottomrule
\end{tabular}
\end{table}

%% file: appendix/setting/pretraining_repa.tex
\begin{table}[h]
\centering
\caption{Pre-training settings for REPA.}
\label{tab:pretrain_repa_full}
\setlength{\tabcolsep}{12pt}
\renewcommand{\arraystretch}{1.05}
\begin{tabular}{l l} 
\toprule
\textbf{Config} & \textbf{Value} \\
\midrule
Optimizer & AdamW \\
Base learning rate & $1\times10^{-4}$ \\
Weight decay & 0.02 \\
Momentum & $(0.9, 0.95)$ \\
Batch size & 2048 \\
LR schedule & Constant \\
Warmup epochs & 12 \\
Training epochs & 49 \\
Gradient clip & 3.0 \\
Label dropout & 0.1 \\
Augmentation & Random Horizontal Flipping \\
\midrule
Teacher encoder & CLIP-L \\
Masking ratio (min) & 0.7 \\
Masking ratio (max) & 1.0 \\
Masking ratio (std) & 0.25 \\
Weight on REPA loss & 0.5 \\
\bottomrule
\end{tabular}
\end{table}

%% file: appendix/setting/pretraining_dream.tex
\begin{table}[h]
\centering
\caption{Pre-training settings for DREAM.}
\label{tab:pretrain_dream_full}
\setlength{\tabcolsep}{12pt}
\renewcommand{\arraystretch}{1.05}
\begin{tabular}{l l} 
\toprule
\textbf{Config} & \textbf{Value} \\
\midrule
Optimizer & AdamW \\
Base learning rate & $1\times10^{-4}$ \\
Weight decay & 0.04 \\
Momentum & $(0.9, 0.95)$ \\
Batch size & 2048 \\
LR schedule & Constant \\
Warmup epochs & 12 \\
Training epochs & 49 \\
Gradient clip & 3.0 \\
Label dropout & 0.1 \\
Augmentation & Random Horizontal Flipping \\
\midrule
Masking ratio (min) & 0.0 \\
Masking ratio (max) & 1.0 \\
Masking ratio (std) & 0.55 \\
Masking warmup epochs & 36 \\
Maximum masking for CLIP loss & 75\% \\
Minimum masking for diffusion loss & 50\% \\
Weight on CLIP loss & 0.005 \\
\bottomrule
\end{tabular}
\end{table}

%% file: appendix/setting/linear_probing.tex
\begin{table}[h]
\centering
\caption{Linear Probing Settings.}
\label{tab:linear_probe_settings}
\setlength{\tabcolsep}{12pt}
\renewcommand{\arraystretch}{1.05}
\begin{tabular}{l l} 
\toprule
\textbf{Config} & \textbf{Value} \\
\midrule
Optimizer & LARS~\cite{you2017large} \\
Base learning rate & 0.1 (B), 0.05 (L, H), 0.02 (G) \\
Weight decay & 0 \\
Optimizer momentum & 0.9 \\
Batch size & 4096 \\
Learning rate schedule & Cosine decay~\cite{loshchilov2016sgdr} \\
Warmup epochs & 10 \\
Training epochs & 90 \\
Augmentation & RandomResizedCrop \\
\bottomrule
\end{tabular}
\end{table}

%% file: appendix/setting/finetuning.tex
\begin{table}[h]
\centering
\caption{Fine-tuning Settings.}
\label{tab:finetune_settings}
\setlength{\tabcolsep}{12pt}
\renewcommand{\arraystretch}{1.05}
\begin{tabular}{l l} 
\toprule
\textbf{Config} & \textbf{Value} \\
\midrule
Optimizer & AdamW~\cite{loshchilov2017decoupled} \\
Base learning rate & $2.5 \times 10^{-4}$ \\
Weight decay & 0.05 \\
Optimizer momentum & $\beta_1, \beta_2 = 0.9, 0.999$ \\
Layer-wise LR decay~\cite{bao2021beit} & 0.75 \\
Batch size & 1024 \\
Learning rate schedule & Cosine decay~\cite{loshchilov2016sgdr} \\
Warmup epochs & 5 \\
Training epochs & 50 \\
Label smoothing~\cite{szegedy2016rethinking} & 0.1 \\
Augmentation & RandAug (9, 0.5)~\cite{cubuk2020randaugment} \\
Mixup~\cite{zhang2017mixup} & 0.8 \\
CutMix~\cite{yun2019cutmix} & 1.0 \\
Random erase & 0 \\
Drop path~\cite{huang2016deep} & 0.2 \\
\bottomrule
\end{tabular}
\end{table}

%% file: appendix/pseudocode/dream.tex
\begin{algorithm*}[h!]
\caption{Training Loop for DREAM}
\label{alg:train_clipmar_final}
\begin{algorithmic}[1]
    \Require model $\mathcal{M}$, VAE, CLIP-style text encoder $\mathcal{E}_{\text{clip}}$, conditioning text encoder $\mathcal{E}_{\text{cond}}$, data loader $\mathcal{D}$, coefficient $\lambda_{\text{clip}}$, epoch index $e$
    \Statex

    \For{each batch $(\mathbf{I}, \mathbf{T})$ in $\mathcal{D}$}
        \State $\mathbf{Z} \gets \text{VAE.Encode}(\mathbf{I})$
        \State $\mathbf{t}_{\text{clip}} \gets \mathcal{E}_{\text{clip}}.\text{Encode}(\mathbf{T})$ \Comment{text embeddings for CLIP loss}
        \State $\mathbf{t}_{\text{cond}} \gets \mathcal{E}_{\text{cond}}.\text{Encode}(\mathbf{T})$ \Comment{text embeddings for conditioning}
        \State $\mathbf{m} \gets \textsc{MaskingScheduler}(\mathbf{Z},~\hat{s}=e+\tfrac{i}{|\mathcal{D}|})$ \Comment{mask depends on $\mathbf{Z}$ and training step}
        \State $\mathbf{Z}_{\text{unmask}} \gets \textsc{Unmask}(\mathbf{Z}, \mathbf{m})$ \Comment{retain only unmasked tokens}
        \Statex

        \State $(\mathbf{H}, \mathbf{R}) \gets \textsc{EncoderForward}(\mathbf{Z}_{\text{unmask}})$
        \Comment{(1) forward encoder on unmasked tokens}
        \State $\mathbf{z} \gets \textsc{DecoderForward}(\mathbf{H}, \mathbf{m}, \mathbf{t}_{\text{cond}})$
        \Comment{(2) forward decoder conditioned on text}
        \Statex

        \State $\mathcal{L}_{\text{mar}} \gets \textsc{DiffusionLoss}(\mathbf{z}, \mathbf{Z}, \mathbf{m})$
        \Comment{(3) MAR / diffusion loss}
        \State $\mathbf{f}_{\text{img}} \gets \textsc{MeanPool}(\mathbf{R})$ \Comment{mean-pooled encoder image embedding}
        \State $\mathcal{L}_{\text{clip}} \gets \textsc{CLIPLoss}(\mathbf{f}_{\text{img}}, \mathbf{t}_{\text{clip}})$
        \Comment{(4) image–text CLIP loss}
        \Statex

        \State $\mathcal{L} \gets \mathcal{L}_{\text{mar}} + \lambda_{\text{clip}} \cdot \mathcal{L}_{\text{clip}}$
        \Comment{(5) combined objective}
        \State $\mathcal{L}.\textsc{backprop}()$
        \Comment{(6) backpropagation step}
    \EndFor
    \Statex
\end{algorithmic}
\end{algorithm*}

%% file: appendix/pseudocode/repa.tex
\begin{algorithm*}[h!]
\caption{Training Loop for REPA Baseline, Implemented for Masked Image Modeling}
\label{alg:train_repa_mim}
\begin{algorithmic}[1]
    \Require model $\mathcal{M}$, VAE, conditioning text encoder $\mathcal{E}_{\text{cond}}$, REPA teacher encoder $\mathcal{E}_{\text{teacher}}$, data loader $\mathcal{D}$, coefficient $\lambda_{\text{repa}}$, epoch index $e$
    \Statex

    \For{each batch $(\mathbf{I}, \mathbf{T})$ in $\mathcal{D}$}
        \State $\mathbf{Z} \gets \text{VAE.Encode}(\mathbf{I})$
        \State $\mathbf{t}_{\text{cond}} \gets \mathcal{E}_{\text{cond}}.\text{Encode}(\mathbf{T})$
        \Comment{text embedding used solely for decoder conditioning}
        \State $\mathbf{m} \gets \textsc{MaskingScheduler}(\mathbf{Z},~\hat{s}=e+\tfrac{i}{|\mathcal{D}|})$
        \Comment{mask depends on $\mathbf{Z}$ and training step}
        \State $\mathbf{Z}_{\text{unmask}} \gets \textsc{Unmask}(\mathbf{Z}, \mathbf{m})$
        \Comment{retain only unmasked tokens}
        \Statex

        \State $(\mathbf{H}, \mathbf{R}) \gets \textsc{EncoderForward}(\mathbf{Z}_{\text{unmask}})$
        \Comment{(1) encoder forward on unmasked tokens}
        \State $\mathbf{z} \gets \textsc{DecoderForward}(\mathbf{H}, \mathbf{m}, \mathbf{t}_{\text{cond}})$
        \Comment{(2) decoder reconstructs masked tokens}
        \Statex

        \State $\mathcal{L}_{\text{mar}} \gets \textsc{DiffusionLoss}(\mathbf{z}, \mathbf{Z}, \mathbf{m})$
        \Comment{(3) MAR / diffusion reconstruction loss}
        \Statex

        \State $\mathbf{R}_{\text{teacher}} \gets \mathcal{E}_{\text{teacher}}.\textsc{ForwardFeatures}(\mathbf{I})$
        \Comment{teacher encoder features}
        \State $\mathcal{L}_{\text{repa}} \gets \textsc{REPALoss}(\mathbf{R}, \mathbf{R}_{\text{teacher}})$
        \Comment{(4) REPA encoder alignment loss}
        \Statex

        \State $\mathcal{L} \gets \mathcal{L}_{\text{mar}} + \lambda_{\text{repa}} \cdot \mathcal{L}_{\text{repa}}$
        \Comment{(5) combined MAR + REPA objective}
        \State $\mathcal{L}.\textsc{backprop}()$
        \Comment{(6) backpropagation step}
    \EndFor
    \Statex
\end{algorithmic}
\end{algorithm*}

%% file: appendix/pseudocode/semantically_aligned_decoding.tex
\begin{algorithm*}[t]
\caption{Semantically Aligned Decoding}
\label{alg:minimal_clip_critic}
\begin{algorithmic}[1]
\Require Number of candidates $N$, threshold step $T$, total steps $S$, text $\mathcal{Y}$
\State Initialize candidate states $\{(\mathbf{x}^{(n)}, \mathbf{m}^{(n)})\}_{n=1}^N$
\For{$\text{step} = 0$ to $S{-}1$}
    \If{$\text{step} < T$} \Comment{1) Generate $N$ candidates up to step $T$}
        \For{$n = 1$ to $N$}
            \State $(\mathbf{x}^{(n)}, \mathbf{m}^{(n)}) \leftarrow \text{DecodeStep}(\mathbf{x}^{(n)}, \mathbf{m}^{(n)}, \mathcal{Y})$
        \EndFor
    \ElsIf{$\text{step} = T$} \Comment{2) Score with text, retrieve best, then decode}
        \For{$n = 1$ to $N$}
            \State $s_n \leftarrow \text{CLIPScore}(\mathbf{x}^{(n)}, \mathcal{Y})$
        \EndFor
        \State $n^{*} \leftarrow \arg\max_n s_n$
        \State $(\mathbf{x}, \mathbf{m}) \leftarrow (\mathbf{x}^{(n^{*})}, \mathbf{m}^{(n^{*})})$
        \State $(\mathbf{x}, \mathbf{m}) \leftarrow \text{DecodeStep}(\mathbf{x}, \mathbf{m}, \mathcal{Y})$
    \Else \Comment{3) Continue decoding best candidate}
        \State $(\mathbf{x}, \mathbf{m}) \leftarrow \text{DecodeStep}(\mathbf{x}, \mathbf{m}, \mathcal{Y})$
    \EndIf
\EndFor
\State \Return $\text{Unpatchify}(\mathbf{x})$
\end{algorithmic}
\end{algorithm*}